\documentclass[lettersize,journal]{IEEEtran}
\usepackage{amsmath,amsfonts}
\usepackage{array}
\usepackage[caption=false,font=normalsize,labelfont=sf,textfont=sf]{subfig}
\usepackage{textcomp}
\usepackage{stfloats}
\usepackage{url}
\usepackage{verbatim}
\usepackage{graphicx}
\usepackage{cite}
\usepackage{multirow}
\usepackage{bm}
\usepackage{booktabs}
\usepackage[colorlinks=true,linkcolor=black,hidelinks]{hyperref}

\usepackage{subfig} 
\usepackage{subfloat}
\makeatother

\usepackage[linesnumbered,ruled,vlined]{algorithm2e}
\usepackage{algpseudocode}  
\usepackage{amsmath}  

\begin{document}

\title{NAMR-RRT: Neural Adaptive Motion Planning for Mobile Robots in Dynamic Environments}

\author{Zhirui Sun, Bingyi Xia, Peijia Xie, Xiaoxiao Li and Jiankun Wang, \emph{Senior Member, IEEE} 
\thanks{This work is partially supported by National Natural Science Foundation of China Grant \#62103181, Shenzhen Science and Technology Program under Grant RCBS20221008093305007, 20231115141459001, Young Elite Scientists Sponsorship Program by CAST under Grant 2023QNRC001, High level of special funds (G03034K003) from Southern University of Science and Technology, Shenzhen, China. \emph{(Corresponding author: Jiankun Wang).}}
\thanks{Zhirui Sun, Bingyi Xia, Peijia Xie and Jiankun Wang are with Shenzhen Key Laboratory of Robotics Perception and Intelligence, Department of Electronic and Electrical Engineering, Southern University of Science and Technology, Shenzhen, China (e-mail: \url{sunzr2023@mail.sustech.edu.cn}; \url{xiaby2020@mail.sustech.edu.cn}; \url{xiepj2022@mail.sustech.edu.cn}; \url{wangjk@sustech.edu.cn}).}%
\thanks{Zhirui Sun and Jiankun Wang are also with Jiaxing Research Institute, Southern University of Science and Technology, Jiaxing, China.}%
\thanks{Bingyi Xia is also with the Peng Cheng National Laboratory, Shenzhen, China.}%
\thanks{Xiaoxiao Li is with the Harbin Institute of Technology Shenzhen, Shenzhen, China (e-mails: \url{lxx@stu.hit.edu.cn}).}%
}



\maketitle

\begin{abstract} 
Robots are increasingly deployed in dynamic and crowded environments, such as urban areas and shopping malls, where efficient and robust navigation is crucial. Traditional risk-based motion planning algorithms face challenges in such scenarios due to the lack of a well-defined search region, leading to inefficient exploration in irrelevant areas. While bi-directional and multi-directional search strategies can improve efficiency, they still result in significant unnecessary exploration. This article introduces the Neural Adaptive Multi-directional Risk-based Rapidly-exploring Random Tree (NAMR-RRT) to address these limitations. NAMR-RRT integrates neural network-generated heuristic regions to dynamically guide the exploration process, continuously refining the heuristic region and sampling rates during the planning process. This adaptive feature significantly enhances performance compared to neural-based methods with fixed heuristic regions and sampling rates. NAMR-RRT improves planning efficiency, reduces trajectory length, and ensures higher success by focusing the search on promising areas and continuously adjusting to environments. The experiment results from both simulations and real-world applications demonstrate the robustness and effectiveness of our proposed method in navigating dynamic environments. A website about this work is available at \href{https://sites.google.com/view/namr-rrt}{https://sites.google.com/view/namr-rrt}.
\end{abstract}

\def\abstractname{Note to Practitioners}
\begin{abstract}
The growing demand for autonomous robots to navigate efficiently and robustly in dynamic, crowded environments like public areas has motivated this work. Traditional risk-based motion planning algorithms often suffer from unfocused search processes, leading to inefficient exploration and performance bottlenecks. This article introduces the NAMR-RRT algorithm to address these issues by integrating neural network-generated heuristic regions to guide the search process. NAMR-RRT adaptively updates both the heuristic region and sampling rate during the planning process, allowing it to focus on more promising areas and dynamically adjust to environmental changes. Unlike conventional methods relying on random exploration, NAMR-RRT improves efficiency by focusing searches in regions more likely to lead to the feasible path, thereby reducing trajectory length and enhancing overall performance. This approach is valuable for mobile robots operating in human-robot coexisting environments, where dynamic adaptability and efficient navigation are critical. The experiment results demonstrate that NAMR-RRT provides a reliable and efficient solution for motion planning in such complex scenarios. 
\end{abstract}


\section{Introduction}
\IEEEPARstart{I}{n} recent years, autonomous robots have become an integral part of daily life, with their presence expanding across diverse sectors—from automated guided vehicles (AGVs) \cite{agv} in warehouses to cleaning robots \cite{cleaning} in shopping malls. As these robots take on increasingly complex tasks and interact with their surroundings, they face the challenge of navigating unpredictable environments filled with dynamic obstacles and continuously changing conditions. Effective motion planning algorithms are essential to enable autonomous robots to navigate these environments efficiently and robustly.

Over the past few decades, various motion planning algorithms have been proposed, each with its characteristics. Grid-based methods, such as A* \cite{A*}, and Dijkstra's \cite{Dijkstra} algorithms, are widely regarded for their completeness and optimality in static, well-structured environments. However, as the dimensionality of the search space increases, these algorithms become computationally expensive and are often unsuitable for real-time planning in dynamic scenarios. Potential field methods, such as Artificial Potential Field (APF) \cite{apf}, which generate motion by treating the robot as a particle influenced by attractive forces towards the goal and repulsive forces from obstacles, offer faster computation but often suffer from local minima, causing the robot to get stuck before reaching the target. Optimization-based methods, such as trajectory smoothness \cite{smothness} and trajectory generation \cite{traj-generate}, are effective in producing smooth, optimal trajectories but are highly sensitive to parameter tuning and environment changes. In addition, learning-based approaches like Deep Reinforcement Learning (DRL) \cite{rl} have attracted attention for their ability to train models that predict feasible actions. Despite this, they often lack interpretability, as their black-box nature makes it challenging to understand and predict their decision-making processes.
Sampling-based methods, such as Probabilistic Roadmap (PRM) \cite{PRM} and Rapidly-exploring Random Tree (RRT) \cite{RRT}, are particularly effective in navigating high-dimensional state spaces and incorporating multiple constraints. PRM constructs a roadmap by randomly sampling the state space and connecting feasible points, making it efficient for static environments. However, its need for preprocessing limits its effectiveness in changing scenarios. In contrast, RRT incrementally builds a tree from the start node towards the goal, quickly covering large areas of the state space without a predefined roadmap. This feature makes RRT well-suited for navigating complex and changing environments, providing a foundation for various extensions and improvements in motion planning research.

Various approaches have been proposed to improve the performance of RRT, such as RRT-Connect \cite{rrt-connect} and RRdT* \cite{rrdt}. These methods improve planning efficiency by growing trees from the start and goal points or using multiple search directions. However, due to the nonholonomic constraints of robots \cite{constrain}, they often encounter the Two-Point Boundary Value Problem (TBVP) \cite{tbvp}, which prevents the direct connection of nodes between two trees. Solving TBVP is computationally expensive, and solutions are not always guaranteed. To address this challenge, researchers introduce heuristic-based search methods that bypass the need to solve TBVP directly. It leads to developing algorithms such as B2U-RRT \cite{b2u} and MT-RRT \cite{mt-rrt}, which improve the search process without addressing TBVP explicitly. However, these methods still do not consider dynamic environments.
Risk-RRT \cite{risk-rrt} is introduced as a method to handle motion planning and obstacle avoidance in dynamic environments by incorporating risk awareness into the search process. Building upon this, further improvements such as Bi-Risk-RRT \cite{bi-risk-rrt} and Multi-Risk-RRT \cite{Multi-Risk-RRT} enhance performance by incorporating bi-directional and multi-directional search strategies. These methods enable more efficient robot navigation in dynamic environments. However, despite these enhancements, they still have limitations, especially without a clearly defined search region. It leads to inefficient exploration, as significant computational resources are often wasted in irrelevant areas.

\begin{figure}[t]
\centering
    \includegraphics[width=1\columnwidth]{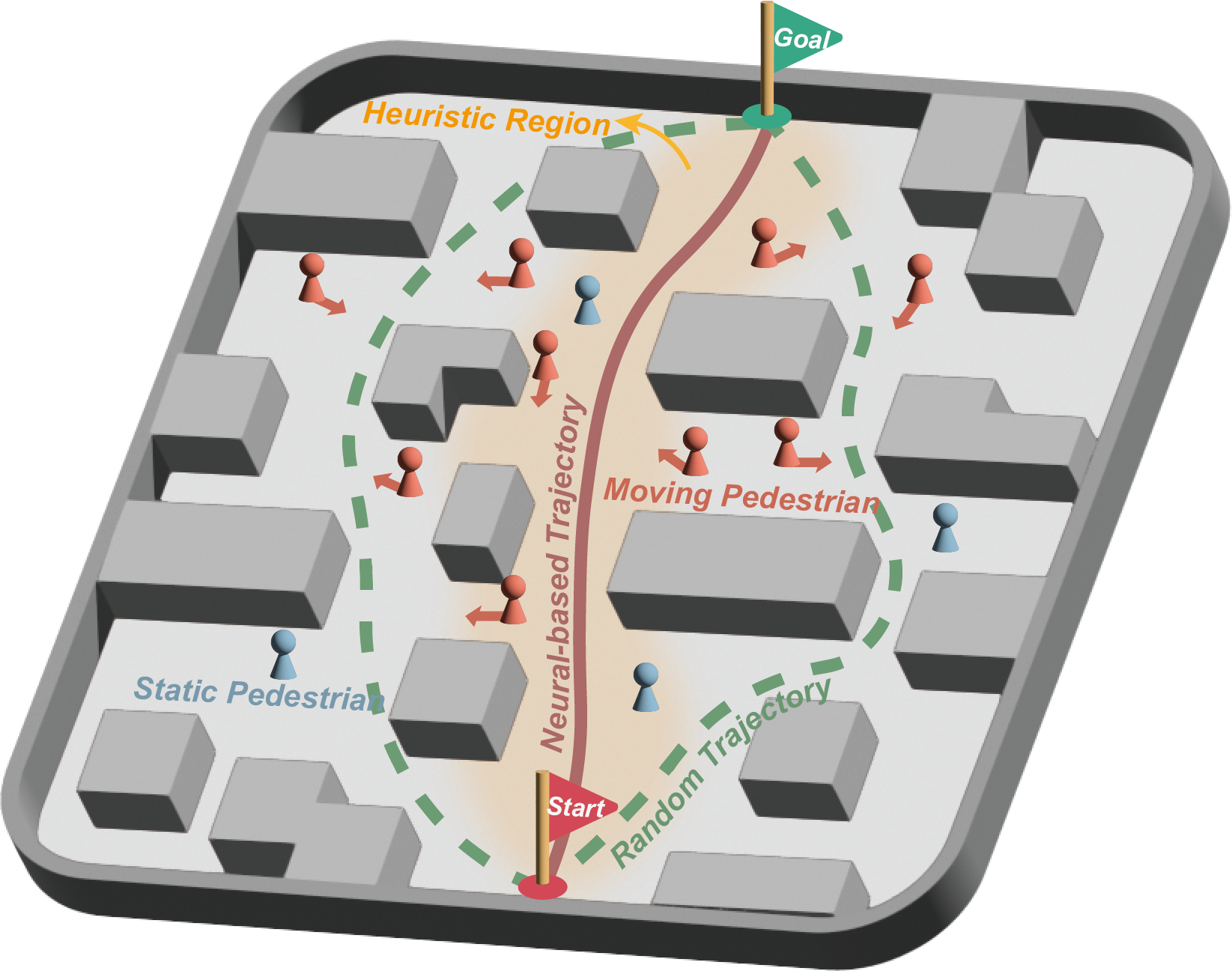}
\caption{The diagram of the robot's navigation in a dynamic environment. The robot starts at the red flag and moves towards the green flag. Static and moving pedestrians are shown as blue and red icons. The Heuristic Region (yellow) guides the robot's search. The Neural-based Trajectory (red) highlights the efficient trajectory guided by this region, while the Random Trajectory (green) represents an inefficient trajectory from random sampling.}
\label{top}
\end{figure}

This article presents a novel algorithm called NAMR-RRT to address the challenges mentioned above. As shown in Fig. \ref{top}, NAMR-RRT utilizes neural network-generated heuristic regions to guide the search towards more promising areas, reducing the time and computational cost of unnecessary exploration. NAMR-RRT dynamically updates the heuristic region and the sampling rate, allowing it to adapt quickly to environmental changes. By incorporating this adaptive feature, NAMR-RRT enhances planning performance, improving the algorithm's robustness in dynamic environments.
The main contributions of this article are summarized as follows:
\begin{itemize}
    \item This article introduces a neural network model based on PointNet++ that employs parallel inference and iterative generation to create heuristic regions, improving the search process efficiency by directing exploration towards more promising areas.
    \item The proposed NAMR-RRT algorithm integrates Neural Adaptive Guiding, Multi-directional Searching, and Risk-aware Growing, providing a comprehensive solution for efficient navigation in dynamic environments.
    \item The adaptive updating of the heuristic region and sampling rate enables NAMR-RRT to adjust during planning dynamically, enhancing its responsiveness to changing environments. Extensive comparative experiments verify the effectiveness of this feature.
\end{itemize}

\section{Related Work}\label{related-work}
Classical sampling-based motion planning algorithms like RRT \cite{RRT} and PRM \cite{PRM} are widely used in robotic motion planning problems. RRT explores state spaces incrementally by building a tree, while PRM connects sampled points in the environment with collision-free paths. 
For path optimality, RRT* \cite{rrt*} builds on RRT by incrementally refining paths to minimize cost. Informed-RRT* \cite{Informed_rrt} and Batch Informed Trees \cite{bit} improve convergence speed by focusing sampling on regions likely to improve the current best path. 
Methods such as RRT-Connect \cite{rrt-connect} and RRdT* \cite{rrdt} have been introduced to increase planning efficiency. These approaches accelerate the search process by expanding trees from the start and goal points or exploring the state space through multiple search directions. It significantly reduces the time required to find a feasible path. Extensions like B2U-RRT \cite{b2u} and MT-RRT \cite{mt-rrt} take the robot's kinematic constraints into account, further enhancing performance in constrained environments. While effective in static environments, these methods face challenges when applied to dynamic environments.

Motion planning in dynamic environments introduces additional complexity due to the need to account for moving obstacles and continuously changing conditions. Several algorithms have been proposed to address these challenges.
MP-RRT \cite{mprrt} extends RRT for dynamic environments by biasing the sampling distribution and reusing branches from previous iterations. Grid-based methods like D* \cite{d*} and D*-Lite \cite{d*-lite} are designed for dynamic environments. However, the environment's discretization and problem complexity often constrain their effectiveness.
RRTX \cite{rrtx} refines and repairs the same search graph throughout navigation, efficiently updating paths in dynamic environments.
RT-RRT* \cite{re-rrt} incorporates an online tree rewiring strategy that enables the tree root to move with the agent, retaining previously sampled paths rather than discarding them.
Fulgenzi \emph{et al}. \cite{risk-rrt} propose a time-based tree framework incorporating risk awareness into the search process, allowing the algorithm to assess and avoid potential risks from dynamic obstacles. Expanding on this foundation, Bi-Risk-RRT \cite{bi-risk-rrt} and Multi-Risk-RRT \cite{Multi-Risk-RRT} improve planning efficiency and success rates by incorporating bi-directional and multi-directional search strategies, making these algorithms more suitable for complex and dynamic environments. These methods, while effective, still have limitations in terms of focused exploration, as they cannot clearly define and concentrate the search within promising areas.

Recent advances in motion planning have explored integrating machine learning techniques, particularly neural networks, to predict feasible paths or regions for exploration. Neural network-based methods offer the potential for more intelligent planning by leveraging learned environmental knowledge. For example, Neural RRT* \cite{neural-rrt} and Neural Informed RRT* \cite{neural-informed-rrt} apply neural networks to guide the sampling process, improving path exploration and convergence in state spaces. 
Meng \emph{et al}. \cite{tis} use a Deep Invertible Koopman operator with control U (DIKU) to generate time-informed sets (TIS), enhancing non-uniform sampling in kinodynamic motion planning and guiding the search towards more feasible trajectories in nonlinear systems.
Zhang \emph{et al}. \cite{gan} utilize a generative adversarial network (GAN) to create heuristic regions to improve non-uniform sampling, directing the search towards more promising areas in the environment. 
However, many of these methods operate with static heuristic regions, limiting their adaptability in dynamic environments.
In response to these challenges, NAMR-RRT is developed to combine the strengths of neural network-based heuristics with the flexibility of adaptive search strategies. Unlike neural network-based methods with a fixed heuristic region and sampling rate, NAMR-RRT continuously updates the heuristic region and sampling rate, enabling it to adapt to changing environments dynamically. This approach not only improves planning efficiency but also ensures more targeted exploration, addressing the shortcomings of prior work and offering enhanced performance in dynamic and complex scenarios.

The remainder of this article is organized as follows: Section III provides the formulation of motion planning and the time-based RRT. Section IV details the proposed NAMR-RRT algorithm, explaining the integration of Neural Adaptive Guiding, Multi-directional Searching, and Risk-aware Growing Function. Section V covers the details of the implementation, the results of simulation experiments compared with the baseline algorithms, and the deployment of NAMR-RRT in the real-world. Section VI provides an in-depth discussion, focusing on the performance of the baseline algorithms, the advantages of neural network-based approaches, and the impact of adaptive updates in NAMR-RRT. Section VII concludes with a summary and considerations for future work.

\section{Preliminaries}
In this section, the motion planning problem is formulated in Subsection \ref{motion planning}, followed by introducing the time-based RRT in Subsection \ref{time-based}.
\subsection{Motion Planning}\label{motion planning}
In this subsection, the formulation for robot motion planning is presented. The robot operates in a state space denoted as $X \subset \mathbb{R}^d$. Let $X_{obs} \subset X$ represent the space occupied by obstacles, and $X_{free} = X \setminus X_{obs}$ define the collision-free region. Since the state space is time-varying in dynamic environments, the obstacle space and the free space are represented as $X_{obs}(t)$ and $X_{free}(t)$. The start state is denoted as $x_{start}$, while the goal state is $x_{goal}$. Furthermore, the goal region is defined as $X_{goal}(t) = \{x \in X_{free}(t) \mid ||x - x_{goal}|| < \epsilon\}$, where $\epsilon$ is a predefined threshold. The control space of the robot is given by $U \subset \mathbb{R}^b$, and let $\eta$ denote the robot control parameters. 

The motion planning problem is to find a valid trajectory $\sigma: [0, T] \mapsto X_{free}(t)$ such that $\sigma(0) = x_{start}$ and $\sigma(T) \in X_{goal}(T)$. For every $t \in [0, T]$, $\sigma(t) \in X_{free}(t)$. The planned trajectory is executed by determining a series of control $u: [0, T] \mapsto U$, and $\forall t \in [0, T]$:
\begin{equation}
\left\{\begin{array}{l}
\sigma(t + \Delta t)=F(\sigma(t), u(t)) \\
u(t + \Delta t) \in U(u(t), \eta)
\end{array}\right.
\end{equation}
Where $F$ represents the robot's dynamics, and $U$ denotes the set of possible control outputs. These outputs are constrained by both the control input at the previous node and the inherent limitations of the robot's control system. The control at step $t$, denoted as $u(t) \in U$, determines the resulting state $\sigma(t + \Delta t)$ after applying the control to the current state $\sigma(t)$. The time increment is represented by $\Delta t$.

\subsection{Time-based RRT}\label{time-based}
\begin{figure}[htb]
\centering
    \includegraphics[width=1\columnwidth]{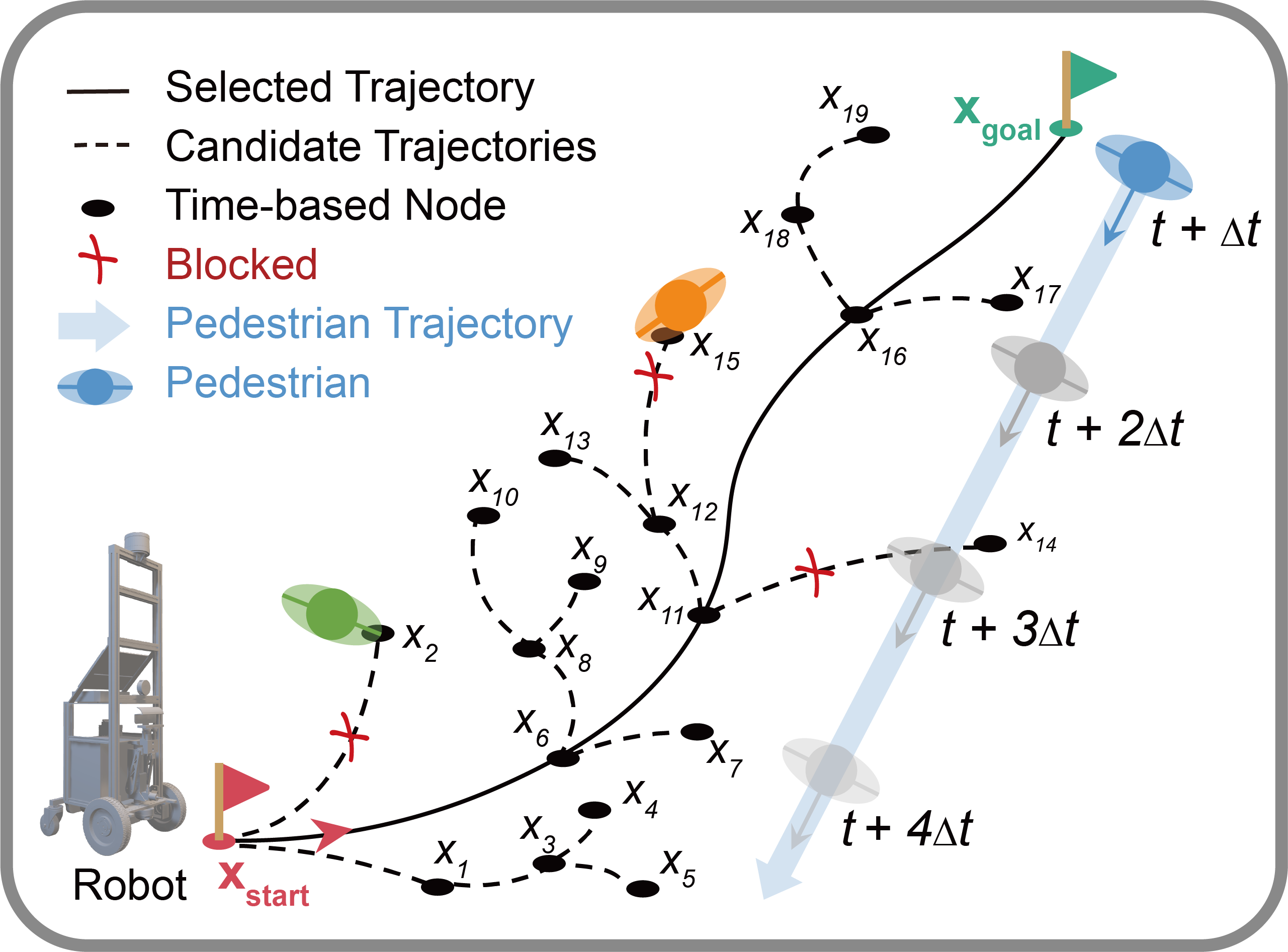}
\caption{The structure of the time-based tree.}
\label{risk}
\end{figure}

The time-based RRT, proposed in \cite{risk-rrt}, enables motion planning in dynamic environments. The structure of the time-based tree is shown in Fig. \ref{risk}. The information associated with each node can be represented by the tuple $(x, x_{parent}, x_{child}, N, t, u, U, P_{collision}(t))$:

\begin{enumerate}
    \item $x$ denotes the robot’s state, including its position and orientation.
    
    \item $x_{parent}$ and $x_{child}$ are the parent and child nodes of the current node. The parent node connects the current node to its upper level in the tree, while the child node corresponds to the nodes generated from it. For example, in Fig. \ref{risk}, $x_6$ is the parent and $x_{16}$ is the child of node $x_{11}$.
    
    \item $N$ represents the depth, which measures the distance between the root node and the current node. The depth increment from a node to its child node is 1. For example, $N(x_{start}) = 0$ and $N(x_7) = 2$.
    
    \item $t$ refers to the timestamp, given by $t = t_0 + n \cdot \Delta t$, where $t_0$ is the timestamp of the $x_{start}$, and $\Delta t$ is the time interval between two nodes.
    
    \item $u=(v,w)$ represents the control input, and $U$ denotes the possible control outputs from the current node to its child node. The robot transitions from the state of its parent node to the current node using $u$
    \begin{equation} 
    x = F(x_{parent}, u).
    \end{equation}
    
    \item At timestamp $t$, the probabilistic collision risk $P_{collision}(t)$  considers both static and dynamic obstacles:
    \begin{equation}
    P_{collision}(t) = P_{static} + (1 - P_{static}) \cdot P_{moving}(t),
    \end{equation}
    where:
    \begin{itemize}
        \item $P_{static}$ represents the probability of collision with static obstacles.
        \item $P_{moving}(t)$ represents the probability of collision with moving obstacles at timestamp $t$.
    \end{itemize}
    Additionally, $P_{moving}(d_k(t))$ indicates the risk from the planned path of a moving obstacle $d_k$ at $t$. The overall collision probability from multiple moving obstacles is calculated as the complement of the product of their non-collision probabilities:
    \begin{equation}
    P_{moving}(t) = 1 - \prod_{k=1}^{m} (1 - P_{moving}(d_k(t))),
    \end{equation}
    where $m$ represents the number of moving obstacles.
\end{enumerate}

To ensure efficient navigation between states, a cost function \cite{risk-dtrrt} is used to quantify the ``effort" required for the robot to move from one state, $x_1$, to another, $x_2$. This function incorporates both the spatial distance and the angular deviation between the two states:
    \begin{equation}
    \textup{Cost}(x_1, x_2)=w_1 || x_1-x_2|| +w_2 \arccos \frac{\overrightarrow{v_1} \cdot \overrightarrow{x_1 x_2}}{\left|\overrightarrow{v_1}\right|\left|\overrightarrow{x_1 x_2}\right|},
    \end{equation}
where $w_1$ and $w_2$ are constants that balance the influence of distance and angle. $\overrightarrow{v_1}$ represents the robot's velocity vector at $x_1$, and $||\cdot||$ is the L2 norm. This cost function evaluates how effectively the robot can transition between states, with lower values indicating more efficient transitions.

In conclusion, the time-based RRT integrates probabilistic collision risk with control inputs to navigate dynamic environments. The cost function further refines this process by ensuring distance and orientation are considered in the robot’s motion planning, optimizing its transitions between states.

\section{NAMR-RRT Algorithm}
\begin{figure*}[htb]
\centering
    \includegraphics[width=2\columnwidth]{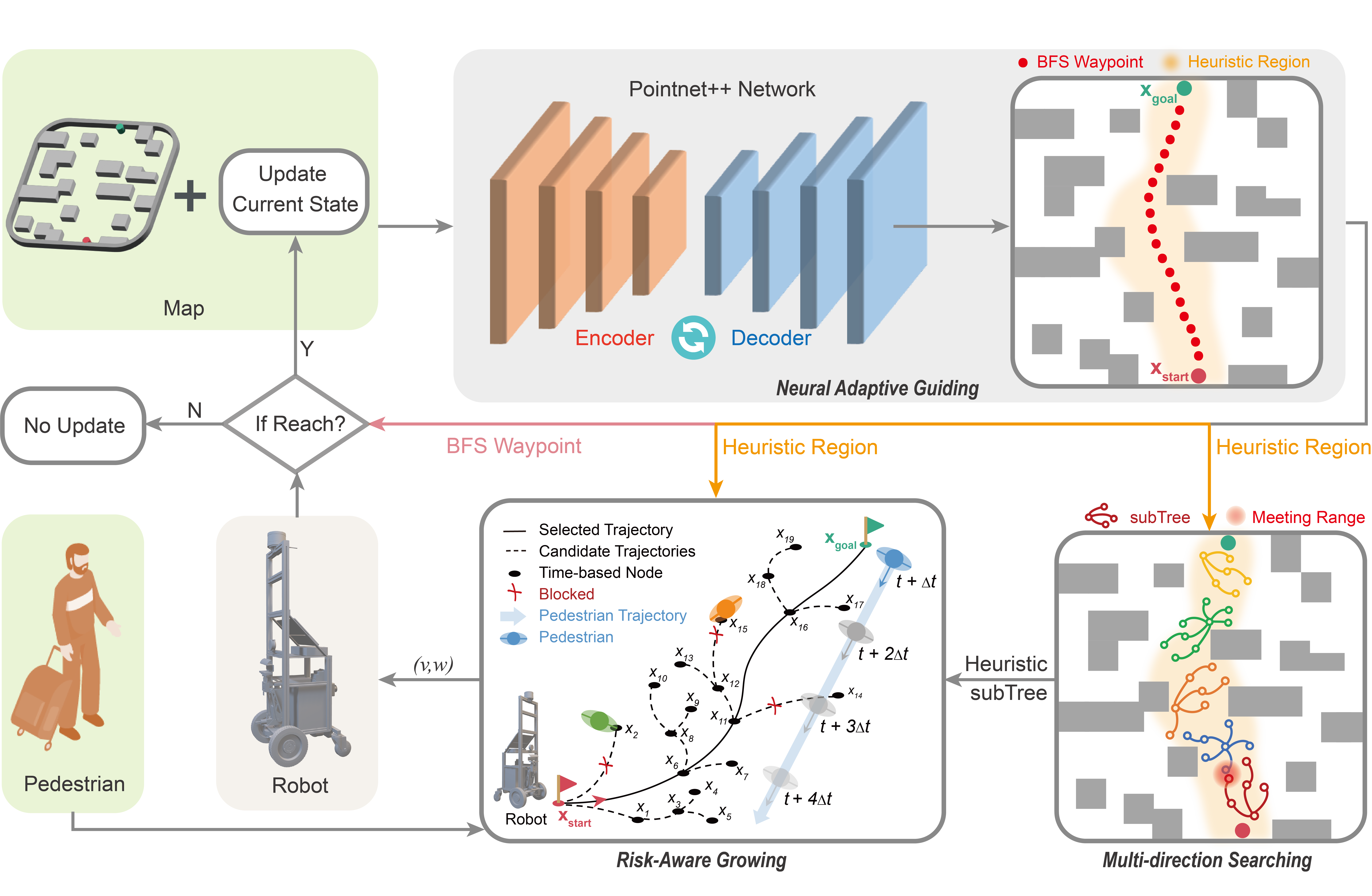}
\caption{A diagram of the architecture of the NAMR-RRT algorithm. The Neural Adaptive Guiding Function first generates the heuristic region based on the input map, followed by a BFS to identify waypoints. The Multi-directional Searching Function then creates subTrees within the heuristic region, while the Risk-aware Growing Function grows the rootTree. The heuristic region and subTrees guide the growth of the rootTree, ultimately generating control instructions for the robot. Additionally, as the robot moves, its position relative to the waypoints is continually assessed, prompting updates to the heuristic region.}
\label{framework}
\end{figure*}
This section presents the architecture of the NAMR-RRT algorithm, as illustrated in Fig. \ref{framework}. The figure outlines the overall framework, highlighting the three core features of the proposed algorithm: Neural Adaptive Guiding, Multi-directional Searching, and Risk-aware Growing.
The input map information is initially fed into a trained neural network model, which infers the heuristic region. A BFS (Breadth First Search) is then performed within this heuristic region to identify waypoints from the start to the goal (Sec. \ref{neural-network}).
Following this, the multi-directional search is executed within the heuristic region, generating subTrees (Sec. \ref{multi-directional}).
Finally, the risk-aware rootTree (Sec. \ref{risk-aware grow}) leverages the heuristic region and subTrees to generate motion instructions for the robot to follow. 
\subsection{NAMR-RRT Algorithm}\label{pseudocode}
\begin{algorithm}  
\caption{NAMR-RRT Algorithm}  
\label{NAMR-RRT}
\KwIn{$x_{start}$, $x_{goal}$, $x_{current}$, $Map$.}
\KwOut{$\mathcal{T}_{opt}$.} 
$closestTree \gets \emptyset$\;
$rootTree \gets \textup{riskGrow}(x_{start})$\; 
$subTrees \gets \textup{multiSearch}(x_{goal}, closestTree)$\; 
\While {$x_{current} \not \in \textup{Region}(x_{goal})$}
{
    \If{$\textup{Meet}(rootTree, subTrees) == \textit{FALSE}$}
    {
    $x_{rand} \gets \textup{neuralSample}(Map)$\;
    
    $closestTree \gets \textup{findClosestTree}(x_{rand})$\;
    
        \If {$closestTree == rootTree$}
        {
            $\textup{riskGrow}(x_{rand})$\;
        }
        \Else
        {
            $\textup{multiSearch}(x_{rand}, closestTree)$\;
        }
    }
    
    \If{$\textup{Meet}(rootTree, subTrees) == \textit{TRUE}$}
    {
       $\textup{subTreeSample}(subTree)$\;
    }
}
\end{algorithm}
In this subsection, the pseudocode of the NAMR-RRT algorithm is presented in Algorithm \ref{NAMR-RRT}. Below is an outline of the algorithm’s main stages:
\begin{enumerate}
    \item Initialization Stage (Lines 1-3): The rootTree is grown using the function \textbf{riskGrow($x_{start}$)}, which builds the main tree while considering the risks from both static and dynamic obstacles. The subTrees are generated through the function \textbf{multiSearch($x_{goal}, closestTree$)}, allowing for exploration of multiple directions.
    
    \item Exploration Stage (Lines 4-11): The algorithm enters a loop until the robot’s current state reaches the goal region. At each step, the function \textbf{Meet($rootTree, subTrees$)} checks whether the rootTree and subTrees are connected. If the trees are not connected, a random state $x_{rand}$ is generated using the function \textbf{neuralSample($Map$)}, leveraging the neural network for guided sampling. The nearest tree to $x_{rand}$ is found using the function \textbf{findClosestTree($x_{rand}$)}. If the closest tree is the rootTree, the function \textbf{riskGrow($x_{rand}$)} expands the rootTree, while if the closest tree is a subTree, the function \textbf{multiSearch($x_{rand}, closestTree$)} expands the subTree.

    \item subTree Guided Stage (Lines 12-13): When the rootTree and subTrees are connected, the function \textbf{subTreeSample($subTree$)} is applied, guiding the rootTree’s growth by sampling from the connected subTree.
\end{enumerate}

\subsection{Neural Adaptive Guiding Function}\label{neural-network}
\begin{algorithm}  
\caption{Neural Adaptive Guiding Function}
\label{adaptive}
\SetKwFunction{neuralSample}{neuralSample}
\SetKwProg{Fn}{Function}{:}{}
\Fn {$\textup{neuralSample}(x_{current}, x_{goal}, Map)$}
{   
$Region, Waypoint \gets \textup{netInfer}(x_{start}, x_{goal}, Map)$\;
$bias \gets 1$\;
    \For{$node \in Waypopint$}
    {
    \If{$\textup{EuD}(x_{current}, node) < \delta$}
    {
    $bias \gets 1$\;
    $nodeList \gets \textup{addNode}(x_{current})$\;
    $lastNode \gets \textup{lastElement}(nodeList)$\;
    $newRegion, newWaypoint, isUpdated \gets \textup{netInfer}(lastNode, x_{goal}, Map)$\;   
    }
    }
    \If{$\textup{Random()} > bias$}
    {
        \If{$isUpdated == TRUE$}
        {
        $Region \gets newRegion$\;
        $Waypoint \gets newWaypoint$\;
        }
    $x_{rand} \gets \textup{randomSample}(Region)$\;
    }
    \Else
    {
    $x_{rand} \gets \textup{randomSample}(Map)$\;
    }
$bias \gets \textup{decayBias}(bias)$\;
    
 $\Return$ $x_{rand}$\;
}
\end{algorithm} 

\begin{algorithm}
\caption{Network Inference Function}  
\label{inference}
\SetKwProg{Fn}{Function}{:}{}
\Fn {$\textup{netInfer}(lastNode, x_{goal}, Map$)}
{
  $Region \gets \emptyset$\;
  $Waypoints \gets \emptyset$\;
  $Iteration \gets 0$\;
  \While{$Iteration < maxIter$ \textbf{and} \textup{BFS}($lastNode$, $x_{goal}$, $Region$) \textbf{fails}}
  {
      $Region \gets \textup{Pointnet++}(lastNode, x_{goal}, Map$)\;
      $Waypoints \gets \textup{BFS}(lastNode, x_{goal}, Region$)\;
      $Iteration++$\;
  
  \If{$Waypoints \neq \emptyset$}
  {
      \Return $Region$, $Waypoints$\;
  }
  \Else
  {
      \Return $Region$, $x_{goal}$\;
  }
  }
  }
\end{algorithm}
The function \textbf{neuralSample} adapts sampling within the region inferred by the neural network, as detailed in Algorithm \ref{adaptive}. The specific details are provided below.
\begin{enumerate}[]
    \item Initialization (Lines 1-3): At first, the function \textbf{netInfer($x_{start}, x_{goal}, Map$)} generates the initial heuristic region and waypoints based on the start state. The bias is initialized to 1, meaning the function begins by greedy sampling in the heuristic region.
    
    \item Heuristic Region and Sampling Rate Updates (Lines 4-9): For each node in the waypoints, the Euclidean distance between the current node and the waypoints is checked. If the distance is less than $\delta$, the current node is added to the $nodeList$, and the last node is used to trigger the function \textbf{netInfer}($lastNode, x_{goal}, Map$) again. The heuristic region is updated, and the bias is reset to 1. It enables more focused sampling within the new region. 
    
    \item Adaptive Sampling (Lines 10-18): If the rootTree fails to grow near the waypoints over several iterations, the bias starts to decay using the function \textbf{decayBias($bias$)}. Random sampling becomes more likely as the bias decreases, allowing the algorithm to explore beyond the heuristic region and expand its search space over time. This decay ensures probabilistic completeness by eventually covering the entire state space.
\end{enumerate}

The function \textbf{netInfer} (Algorithm \ref{inference}) manages neural network inference, ensuring heuristic regions are updated without interrupting the main function. It has two key characteristics. The first is \textbf{Parallel Inference}, where the function \textbf{netInfer} runs simultaneously (Algorithm \ref{adaptive}, Line 9) with the main function. While the main function continues, the heuristic region and sampling rate are only updated after the neural network inference is completed. If the inference is still in progress, the main function continues sampling based on the previous heuristic region. This parallel structure ensures the algorithm does not pause or slow down while waiting for neural network results.

The second feature is \textbf{Iterative Generation}, where the function \textbf{netInfer} performs a BFS search. While generating a heuristic region, the region may not always be fully connected. If BFS fails to find a path from the start node to the goal due to disconnected regions, the neural network is invoked to expand the heuristic region iteratively. This process continues from the point where BFS fails, aiming to generate a connected region over multiple iterations. The number of iterations is capped by a maximum limit ($maxIter$). The algorithm returns the current region and goal node if the valid waypoints is not found after maximum attempts. Through iterative generation, the function strives to obtain a fully connected heuristic region and provide valid waypoints to evaluate the growth of the rootTree in Algorithm \ref{adaptive}. However, even if the neural network does not immediately generate a fully connected region, the rootTree continues to grow, ensuring the process remains uninterrupted.

\subsection{Multi-directional Searching Function}\label{multi-directional}
\begin{algorithm}  
\caption{Multi-directional Searching Function}  
\label{Multi-Tree}
\SetKwProg{Fn}{Function}{:}{}
\Fn {$\textup{multiSearch}(x_{node}, closestTree)$}
{ 
$subTrees \gets \emptyset$\;

    \If {$closestTree \in subTrees$}
    {
        \If{$\textup{Num}(closestTree) == 1$}
        {
        $subTree \gets \textup{addNode}(x_{rand})$\;
        }
        \Else
        {
        $\textup{Merge}(closestTree)$\;
        }
    }
    \Else
    {
        $subTree \gets \textup{randomGenerate}(x_{node})$\;
        $subTrees \gets \textup{addTree}(subTree)$\;
    }
    \Return $subTrees$\;
}
\end{algorithm}

\begin{algorithm}
\caption{subTree-based Sampling Function}
\label{subTree}
\SetKwProg{Fn}{Function}{:}{}
\Fn{\textup{subTreeSample}$(rootTree, subTree, x_{goal})$}
{
        \If{$x_{goal} \in subTree$}{
            $Nodes \gets \textup{extractNode}(subTree, x_{goal})$\;
            \For{$x_{rand} \in Nodes$}{
            $x_{new} \gets \textup{Extend}(rootTree, x_{rand})$\;
            $Count \gets 0$\;
                \While{$x_{new}\notin \textup{Region}(x_{rand})$}
                {
                    $x_{new} \gets \textup{Extend}(rootTree, x_{rand})$\;
                    $Count ++$\;
                    \If{$x_{new}\in \textup{Region}(x_{rand})$}
                    {
                    \If{$x_{rand}==x_{goal}$}
                    {
                     \Return $Reached$\;
                    }
                    \Else
                    {
                        $\textup{Delete}(x_{rand})$\;
                        $\textup{Brake}()$\; 
                    }}
                    \ElseIf{$Count \geq \omega$} 
                    {
                        $\textup{Delete}(x_{rand})$\;
                        $\textup{Brake}()$\; 
                    }
                    }}}
        \Else
        {
            \For{$x_{rand} \in subTree$}
            {
               $x_{new} \gets \textup{Extend}(rootTree, x_{rand})$\;
               \If{$x_{new}\in \textup{Region}(x_{goal})$}
               {
               \Return $Reached$\;
               }
               \Else
               {
               $\textup{Delete}(x_{rand})$\;
               }
}}}
\end{algorithm}
The function \textbf{multiSearch} (Algorithm \ref{Multi-Tree}) manages multiple subTrees that provide heuristic guidance for the rootTree's growth from the start point towards the goal. This function dynamically merges subTrees when they grow close to each other, improving exploration efficiency. The process begins by initializing a subTree at the goal. When a new point $x_{rand}$ is sampled, the algorithm checks if the point is closer to any existing subTree. Depending on the proximity of the sampled point to existing subTrees, the following actions are taken:
\begin{enumerate}
    \item Single subTree Nearby: If the closest subTree is only one, the point is added to that subTree (Lines 4-5).
    \item Multiple subTrees Nearby: If the closest subTrees are more than one, the function \textbf{Merge($closestTree$)} merges these subTrees (Lines 6-7). Merging helps consolidate information and ensures the search becomes more efficient.
    \item No subTree Nearby: If the point is not close to any subTree, a new subTree is generated at that location (Lines 8-10), allowing the search to explore new regions efficiently.
\end{enumerate}
When the rootTree grows near a subTree, the subTree provides heuristic guidance to extend the rootTree using the function \textbf{subTreeSample} (Algorithm \ref{subTree}), as detailed below:
\begin{enumerate}
    \item subTree With Goal: If the subTree includes the goal, its core nodes are sampled multiple times to guide the rootTree towards the goal (Lines 2-18).
    \item subTree Without Goal: If the subTree does not contain the goal, its nodes are used for one-time sampling, guiding the rootTree towards promising regions (Lines 19-25).
\end{enumerate}


\subsection{Risk-aware Growing Function}\label{risk-aware grow}
\begin{algorithm}  
\label{risk-aware}
\caption{Risk-aware Growing Function}  
\SetKwProg{Fn}{Function}{:}{}
\Fn {$\textup{riskGrow}(x_{current}, Map)$}
{   
    $\mathcal{T}_{opt} \gets \emptyset$\;
    $\mathcal{S} \gets \emptyset$\;
    $x_{current} \gets x_{start}$\;
    $time \gets \textup{getCurrentTime}()$\;
    \If {$\mathcal{T}_{opt}$ is empty}
    {
        stopRobot();
    }
    \Else
    {
        $x_{current} \gets \textup{getOneNode}(\mathcal{T}_{opt})$\;
    }
    $\textup{Check}(x_{current})$\;
    $\textup{pruneInvalidTraj}(\mathcal{S}, x_{current})$\;
    $\textup{Check}(staticEnv, DynamicCrowds)$\;
    $time \gets \textup{getCurrentTime}()$\;
    $\textup{predictPosition}(DynamicCrowds, time,...,time+N \times \Delta t)$\;
    
    \If {$Env$ has changed}
    {
        $\textup{updateEnv}(Map, \mathcal{S}, x_{current}, dynamicCrowds)$\;
    }
    \While {$\textup{getCurrentTime}() < time + \Delta t$}
    {
      $x_{new} \gets \textup{Extend}(\mathcal{S}, x_{rand})$\;
        \If {$x_{new} \in \textup{Region}(x_{goal})$}
        {
            \Return $Reached$\;
        }
      }
    
    $\mathcal{T}_{opt} \gets \textup{selectBestTraj}(\mathcal{S})$\;
    $time \gets \textup{getCurrentTime}()$\;
    \Return $\mathcal{S}$\;
  } 
\end{algorithm} 
The \textbf{Risk-aware Growing Function} integrates real-time risk assessment into motion planning by continuously evaluating static and dynamic obstacles, ensuring the robot adapts its trajectory to avoid risks and navigate efficiently in changing environments. 
The function operates in several key stages:
\begin{enumerate}
    \item Initialization (Lines 1-5): The function initializes the trajectory set $\mathcal{T}_{opt}$, rootTree $S$, and the start point $x_{start}$ as the current position $x_{current}$. It tracks the current time to handle dynamic changes.
    \item Risk and Obstacle Handling (Lines 6-12): The robot stops if no optimal trajectory is available. Otherwise, the next node to expand is selected, and potential risks are checked using the function \textbf{Check}($x_{current}$). Invalid trajectories are pruned, and the environment is evaluated for static and dynamic obstacles.
    \item Predicting and Updating Environment (Lines 13-16): The function \textbf{predictPosition} predicts the future positions of dynamic obstacles from $time$ to $time + N \times \Delta t$ and updates the environment when changes occur.
    \item Safe Trajectory Expansion (Lines 17-22): While within a safe time margin, the function \textbf{Extend}($S, x_{rand}$) expands the tree by sampling new points. If a point reaches the goal, the process ends. Otherwise, the best trajectory is selected, and time is updated for the next iteration.
\end{enumerate}
By continuously predicting the movements of dynamic obstacles and updating its environment in real-time, the robot can actively adjust its trajectory and ensure safe navigation.



\subsection{NAMR-RRT Search Process}\label{search_process}
\begin{figure}[htb]
\centering
\subfloat[t = 1.3 s]{\includegraphics[scale=0.22]{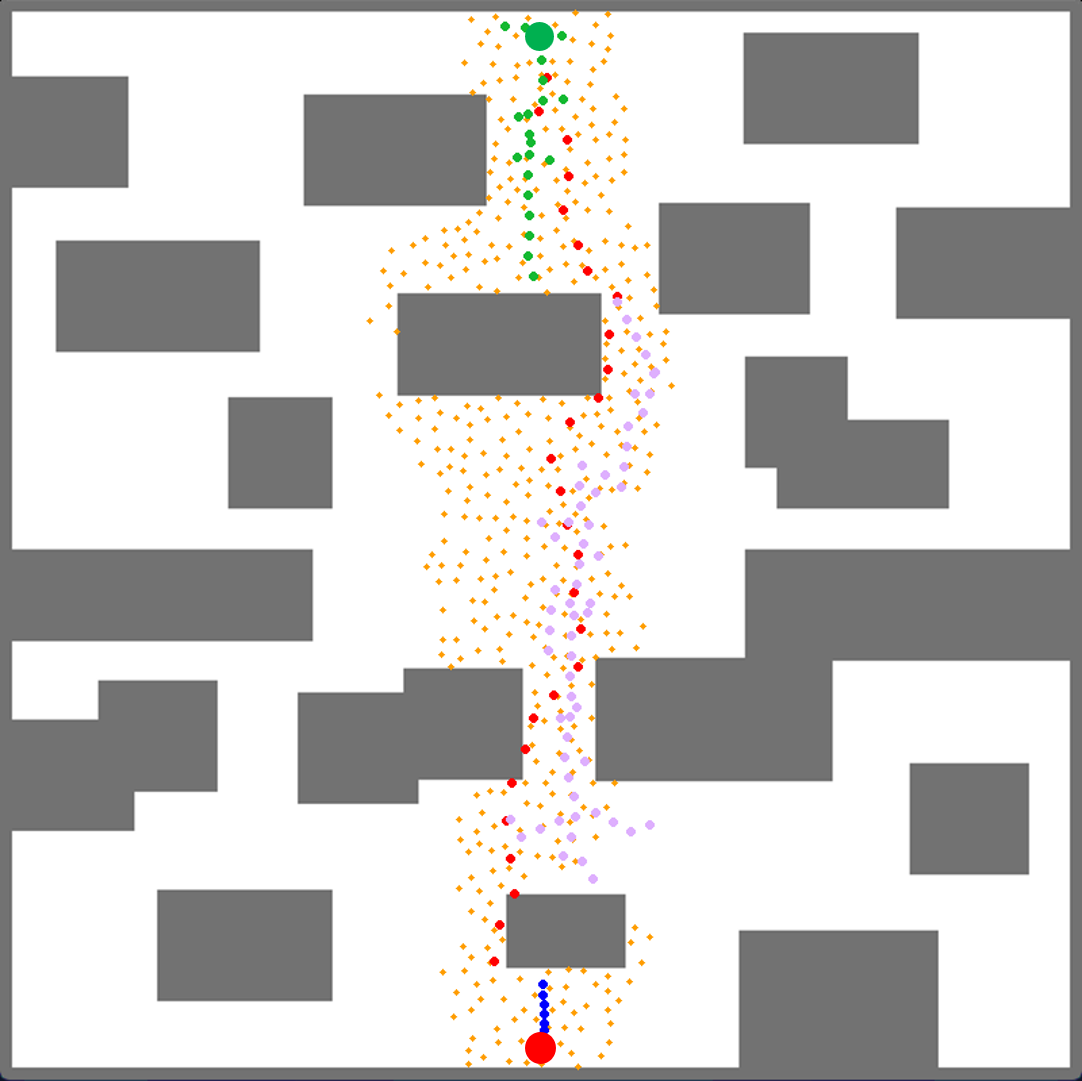}}\vspace{-0.1cm} 
\subfloat[t = 3.9 s]{\includegraphics[scale=0.22]{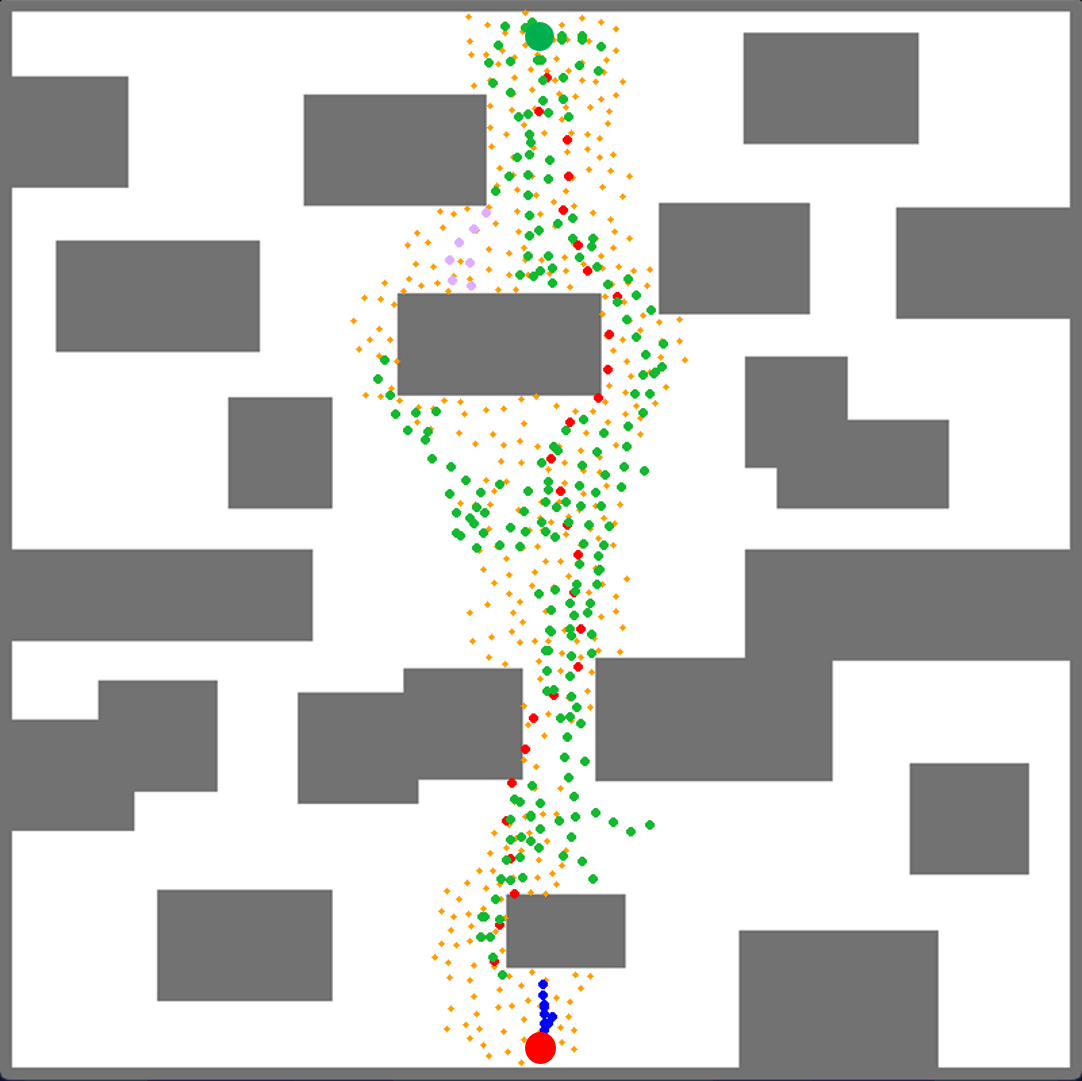}}\vspace{-0.1cm} 
\subfloat[t = 11.9 s]{\includegraphics[scale=0.22]{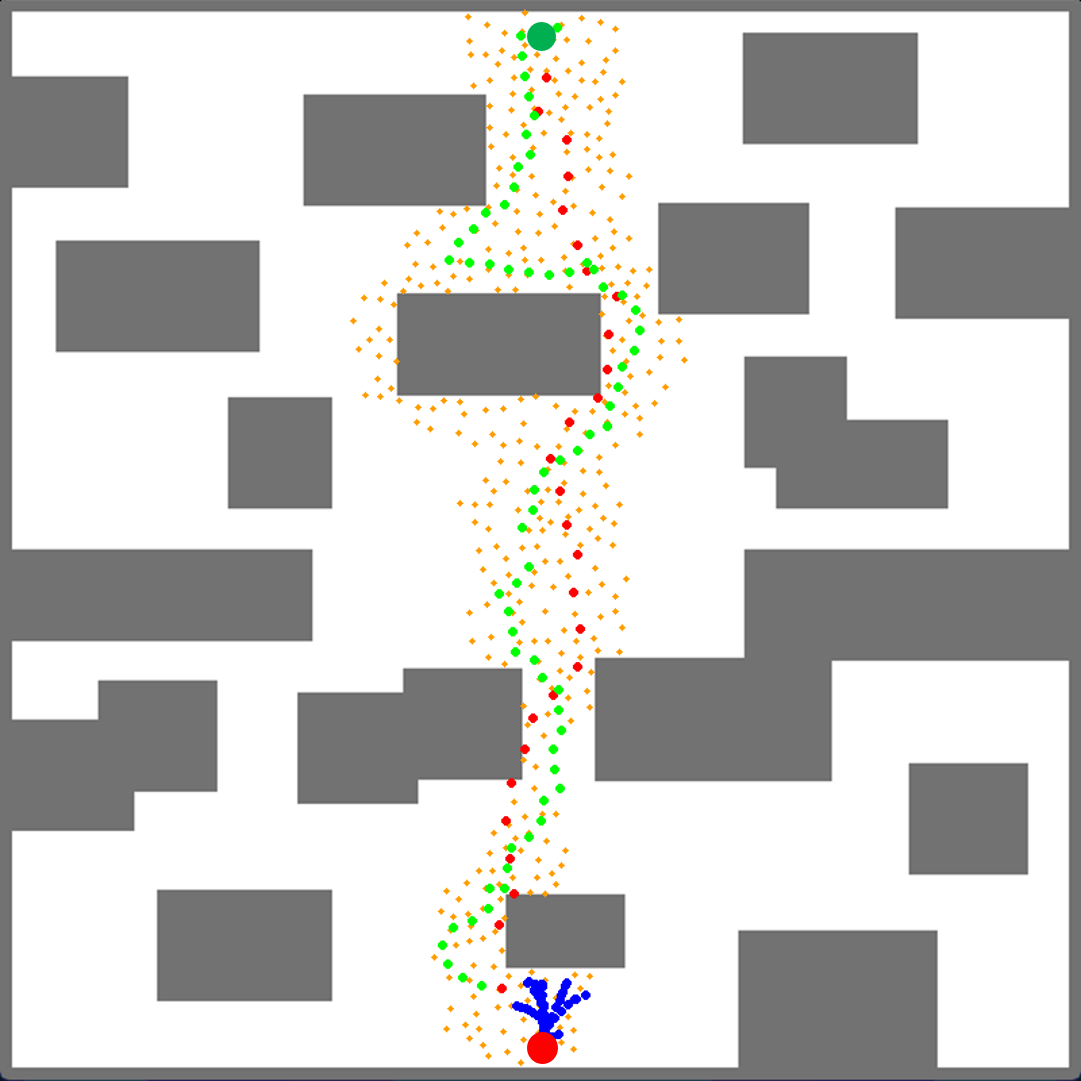}}\vspace{-0.1cm} 
\subfloat[t = 15.1 s]{\includegraphics[scale=0.22]{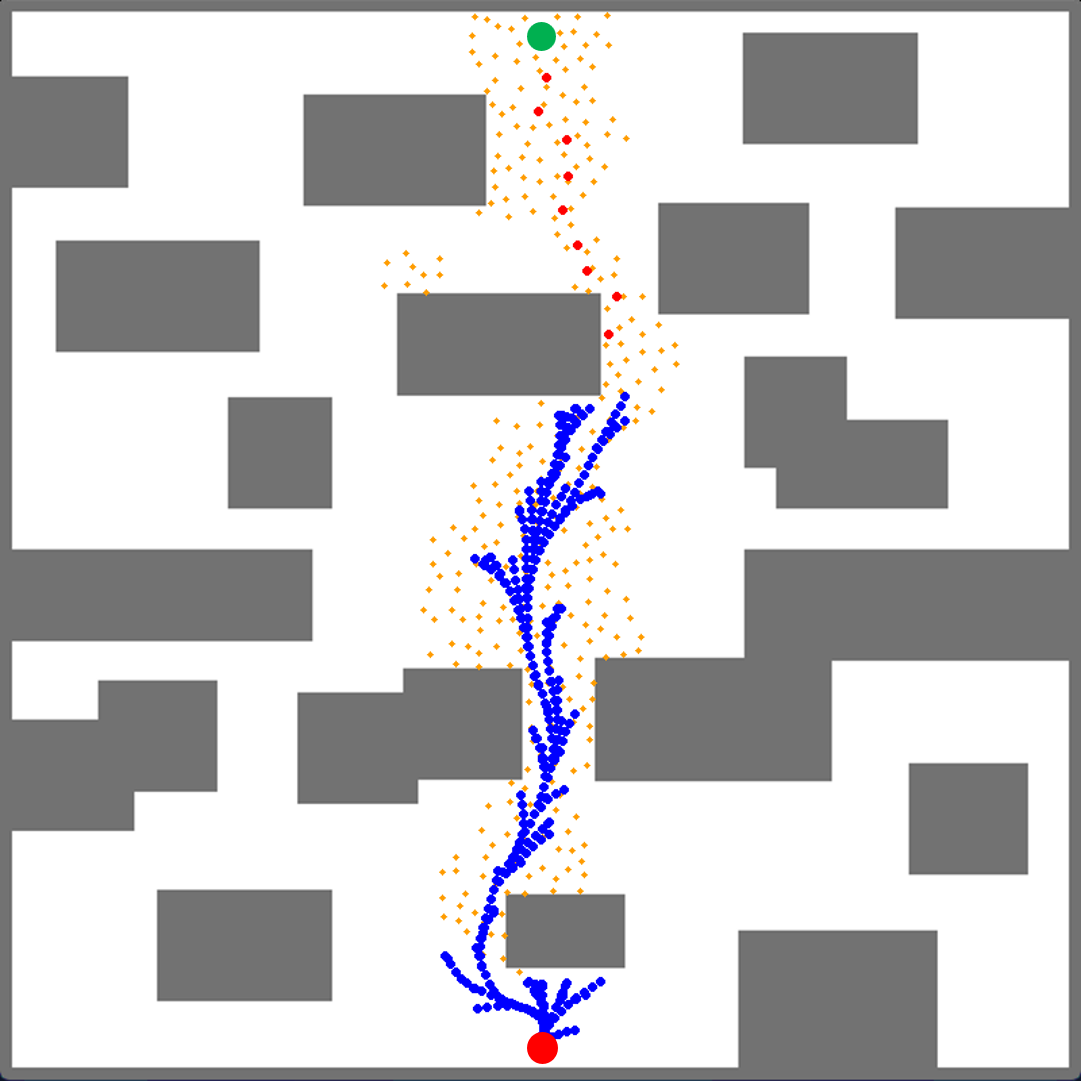}}\vspace{-0.1cm}
\subfloat[t = 19.2 s]{\includegraphics[scale=0.22]{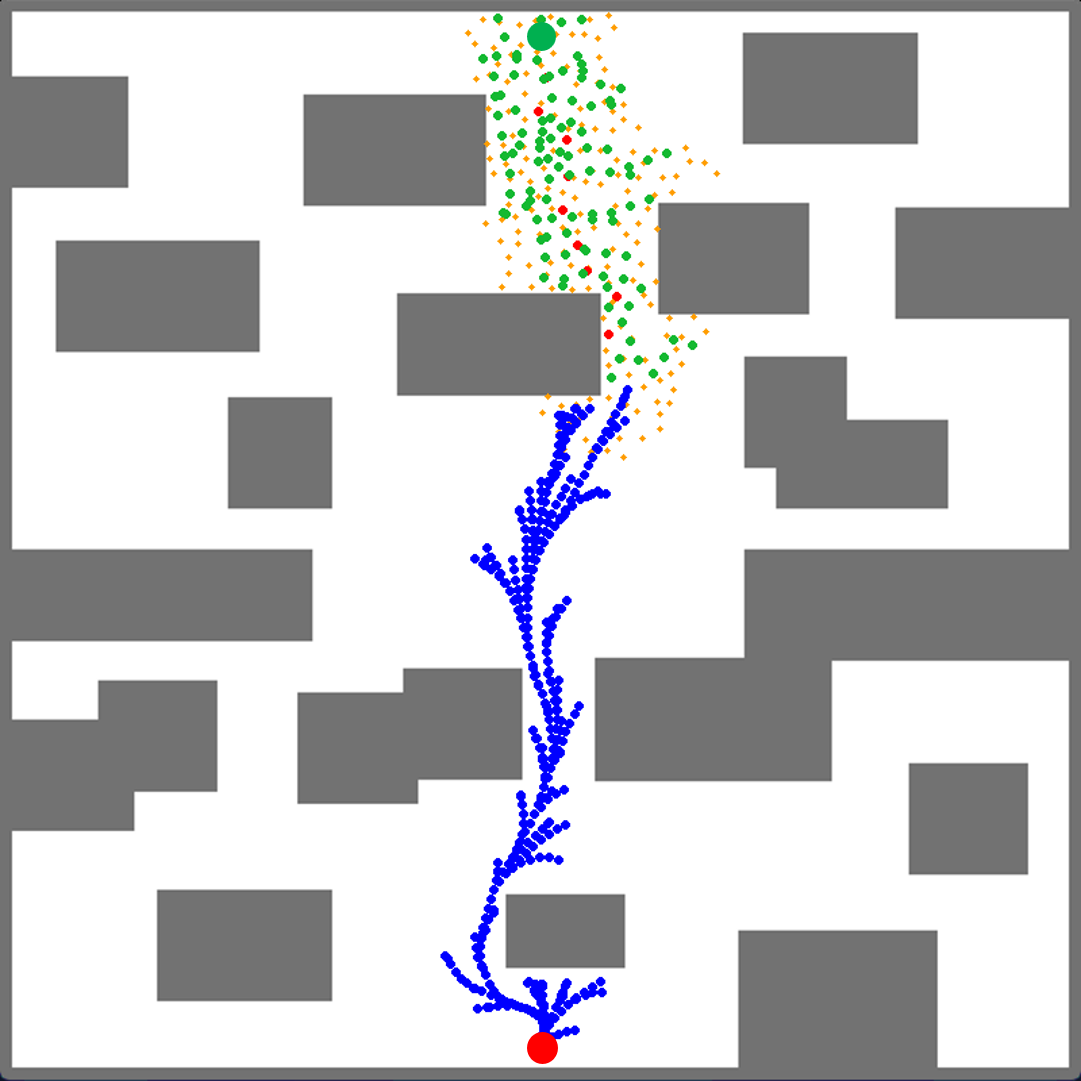}}\vspace{-0.1cm}
\subfloat[t = 20.8 s]{\includegraphics[scale=0.22]{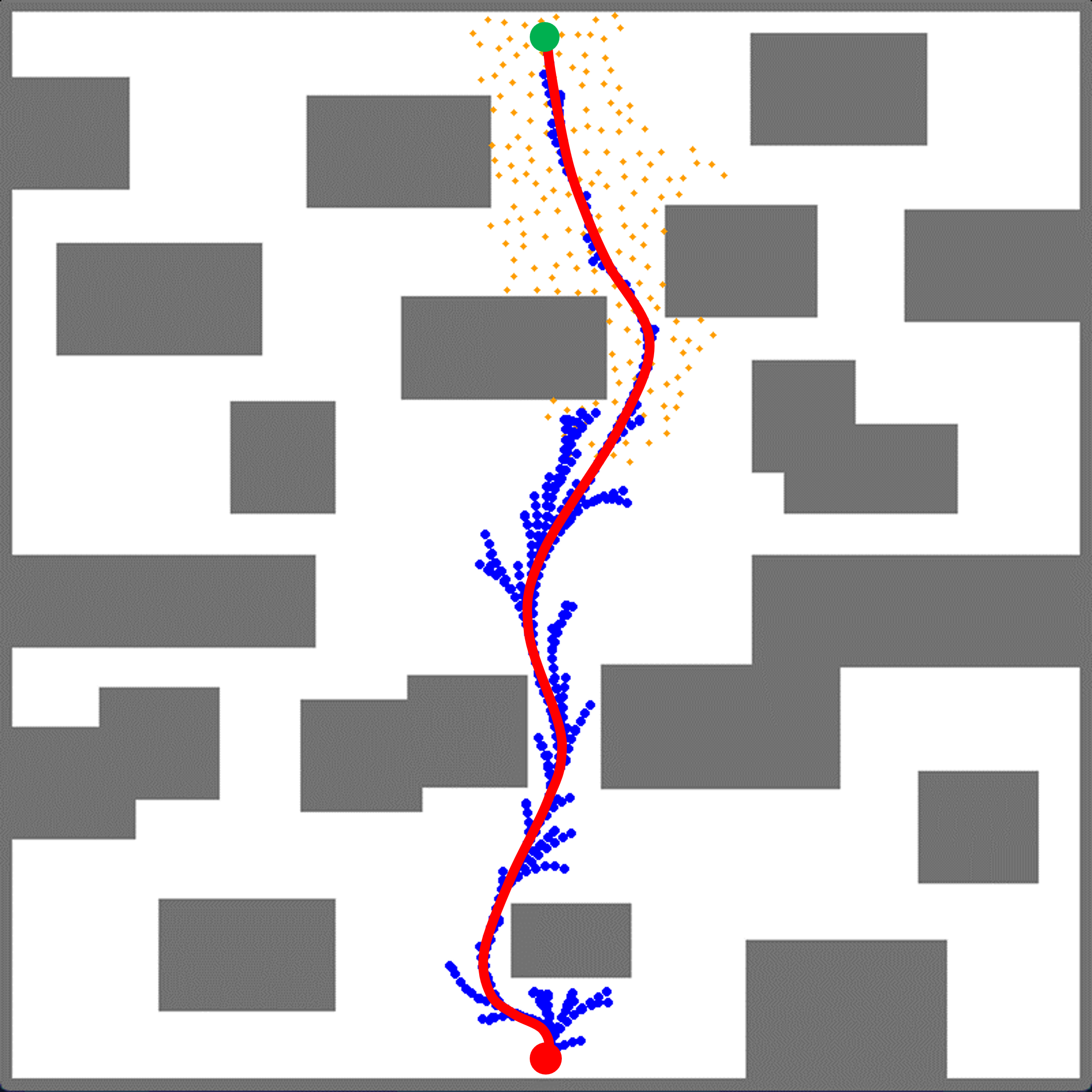}}\vspace{-0.1cm}
\caption{The search process of NAMR-RRT at different key stages. (a) t = 1.3 s, (b) t = 3.9 s, (c) t = 11.9 s, (d) t = 15.1 s, (e) t = 19.2 s, and (f) t = 20.8 s.}
\label{key-stages}
\end{figure}
To comprehensively illustrate the search process of NAMR-RRT, six key stages of its operation on Map-1 are selected, as shown in Fig. \ref{key-stages}. 
\begin{enumerate}
    \item [(a)] Heuristic Searching: At t = 1.3 s, the algorithm begins its multi-directional searching within the heuristic region generated by the neural network (represented by yellow dots). Different colors denote various subTrees, with blue representing the rootTree. The red and green dots mark the start and goal points, respectively, while the small red dots indicate waypoints within the heuristic region.
    \item [(b)] subTree Growing and Merging: At t = 3.9 s, the subTrees have grown and begun to merge into a larger subTree (represented by green), consolidating the information from the different exploration directions. 
    \item [(c)] subTree Near rootTree: At t = 11.9 s, the subTree has grown close to the rootTree and contains the goal point. It triggers the function \textbf{subTreeSample}, where the path is extracted from the goal to the meeting point, and multiple samples are taken along the extracted path (represented in light green) to guide the rootTree's growth towards the goal.
    \item [(d)] rootTree Growing: At t = 15.1 s, the rootTree has grown significantly after being guided by the heuristic subTree.
    The heuristic region has been refined, allowing more concentrated sampling and avoiding unnecessary exploration in irrelevant areas. It ensures the rootTree grows more directly towards the goal.
    \item [(e)] Continued Heuristic Searching: At t = 19.2 s, the algorithm continues multi-directional searching in an updated heuristic region. 
    The continuous updating and shrinking of the heuristic region helped the rootTree focus more efficiently on the goal, concentrating its sampling efforts and avoiding unnecessary spaces. As the sampling rate is updated throughout the process, the rootTree's search becomes more focused, minimizing wasted effort in areas without potential paths.
    \item [(f)] At t = 20.8 s, the rootTree effectively uses the heuristic guidance to grow towards the goal aggressively, reaching the goal point and finding a feasible trajectory (shown in red).
\end{enumerate}

\section{Experiments}\label{experiments}
This section introduces details of the implementation and results of the experiment.
\subsection{Implementation Details}
\subsubsection{Compare Algorithms} NAMR-RRT is compared with several baseline algorithms, including Risk-RRT \cite{risk-rrt}, Bi-Risk-RRT \cite{bi-risk-rrt}, Multi-Risk-RRT \cite{Multi-Risk-RRT}, and NMR-RRT, a variant of NAMR-RRT. NAMR-RRT is the complete version of the algorithm, integrating the neural network-generated heuristic region with dynamic updates to both the region and sampling rate, enabling efficient and adaptive motion planning. NMR-RRT employs a fixed heuristic region and sampling rate without the dynamic adaptation feature.

\subsubsection{Performance Metrics} The performance of the algorithms is evaluated using three metrics: Success Rate, Execution Time, and Trajectory Length. Success Rate represents the ratio of successful attempts to generate the feasible trajectory, where any attempt taking longer than 720 seconds is considered a failure. Execution Time measures how long the robot can reach the goal from the start. Trajectory Length represents the total distance the robot travels from the start to the goal.

\subsubsection{Training Process} PointNet++ \cite{pointnet++} is implemented in this article as the neural network model. Training is performed offline using PyTorch on an AMD EPYC MILAN 7413 CPU and an NVIDIA RTX A6000 GPU. The training setup uses the SDG \cite{sdg} optimizer with an initial learning rate of 0.001, a batch size of 64, and runs for 150 epochs. The dataset consists of 10000 2D random environments, each 300 $\times$ 300 pixels in size. A* generates the optimal path with a step size of one pixel and a clearance of three pixels. A point cloud with 2048 points is generated, and guidance state labels are assigned based on whether each point is within a radius of 10 pixels from the optimal path. 

\subsubsection{Simulation Map} 
\begin{figure}[htb]
\centering
    \includegraphics[width=1\columnwidth]{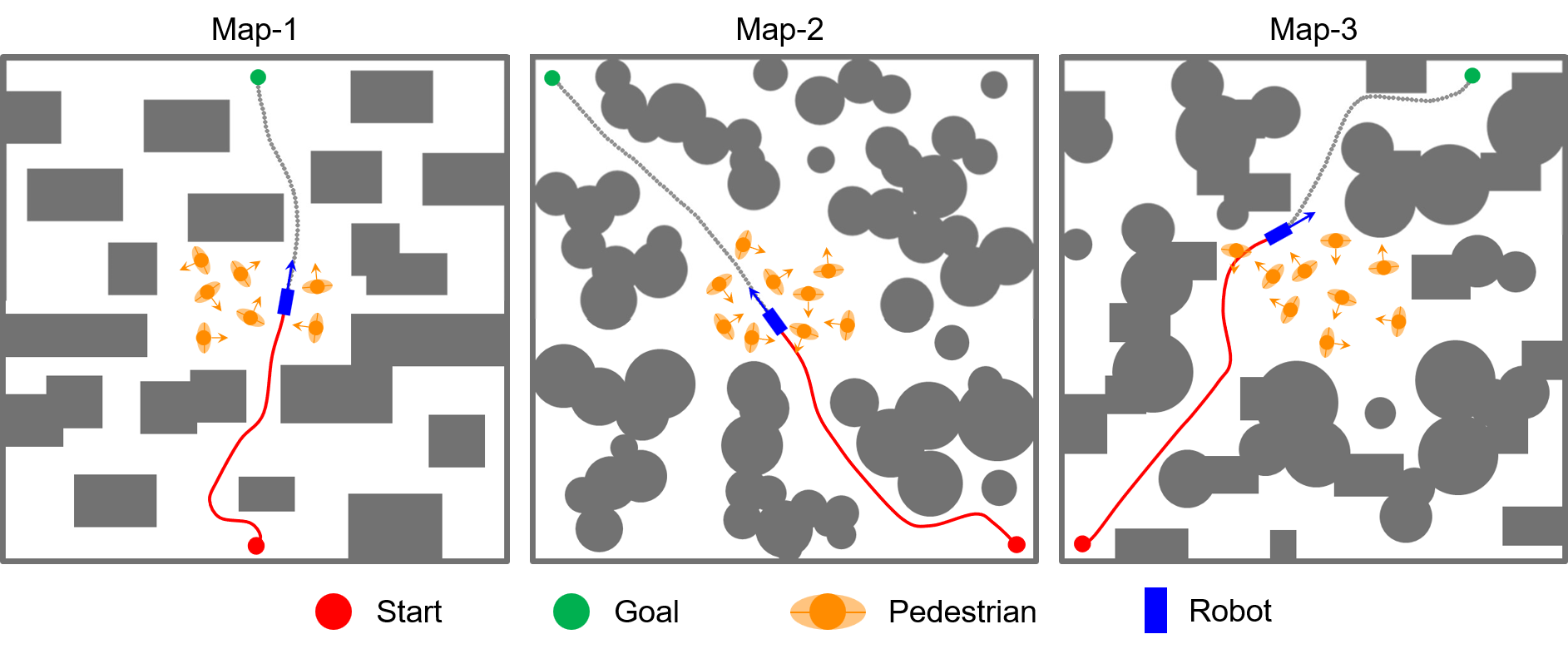}
\caption{Three simulation maps. In each map, the robot is represented by the blue rectangle from the start point (red dot) to the goal point(green dot). The environment contains static obstacles, illustrated as gray blocks, while dynamic obstacles, simulating crowd movement, are represented by orange icons.}
\label{map}
\end{figure}
Fig. \ref{map} presents three simulation maps to evaluate algorithm performance across diverse environments. Map-1 features grid-arranged block obstacles, simulating shelves or stacked goods in a warehouse. Map-2 contains scattered circular obstacles, representing crowds or dispersed objects in markets or malls. Map-3 combines block and circular obstacles, simulating urban environments with buildings and pedestrians.

\subsubsection{Dynamic Environment} 
\begin{figure}[htb]
\centering
    \includegraphics[width=1\columnwidth]{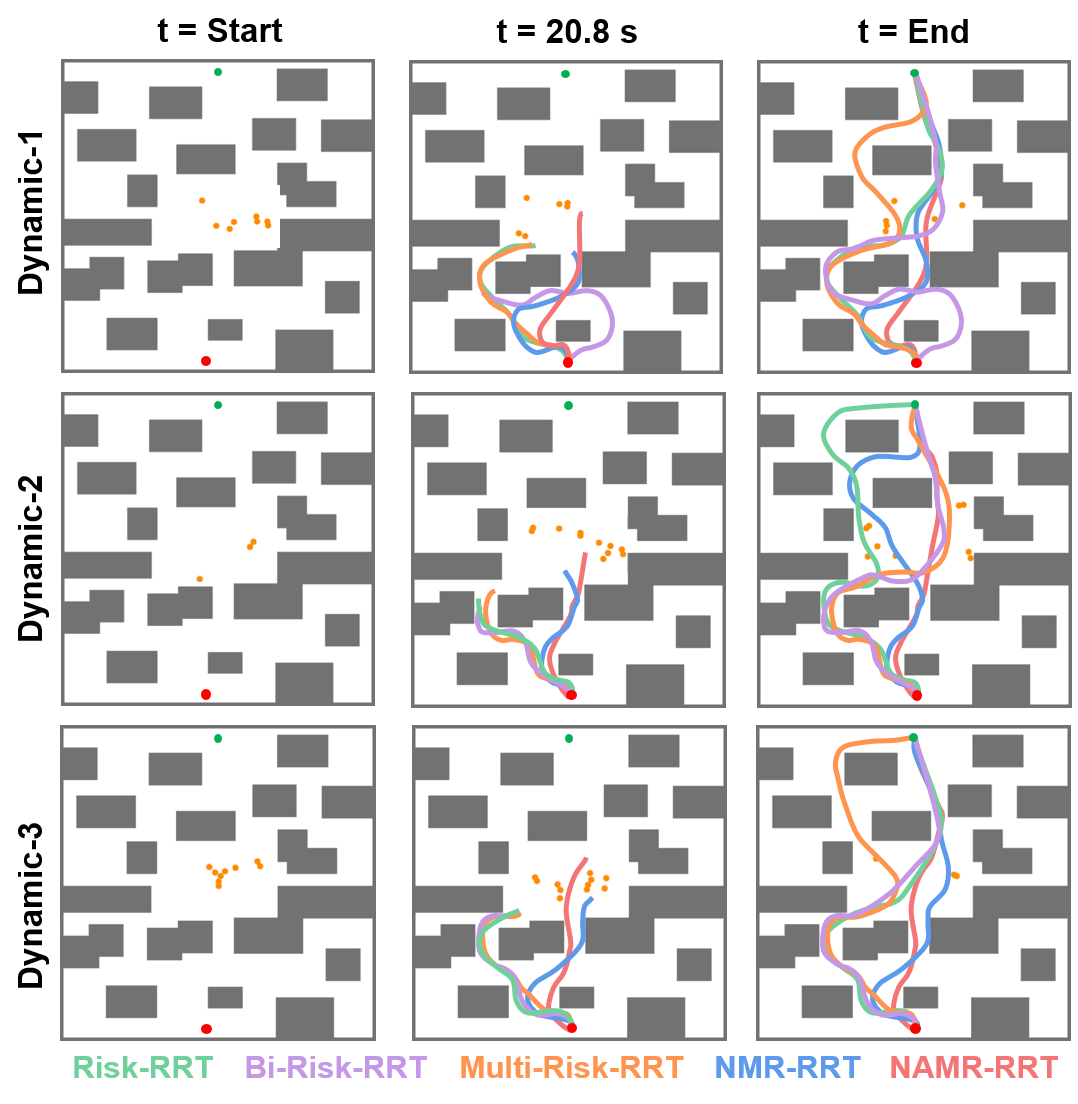}
\caption{Five motion planning algorithms in three dynamic environments. The red and green dots represent the start and goal points, while the orange dots indicate dynamic pedestrians. The trajectories are color-coded by the planning algorithm: green for Risk-RRT, purple for Bi-Risk-RRT, orange for Multi-Risk-RRT, blue for NMR-RRT, and pink for NAMR-RRT.}
\label{trajectory}
\end{figure}
Fig. \ref{trajectory} illustrates the robot's navigation through dynamic environments using different motion planning algorithms, all demonstrated within Map-1. The rows correspond to three dynamic conditions: Dynamic-1, Dynamic-2, and Dynamic-3, with dynamic trajectories from the open-source real-world dataset \cite{ucy}. The figure presents the trajectory planning results of five algorithms—Risk-RRT, Bi-Risk-RRT, Multi-Risk-RRT, NMR-RRT, and NAMR-RRT—at three key stages: the start, 20.8 s, and the end. The trajectories demonstrate how each algorithm adapts to moving pedestrians while navigating around static obstacles.

\subsubsection{Experiment Platform} 
\begin{figure}[htb]
\centering
    \includegraphics[width=0.8\columnwidth]{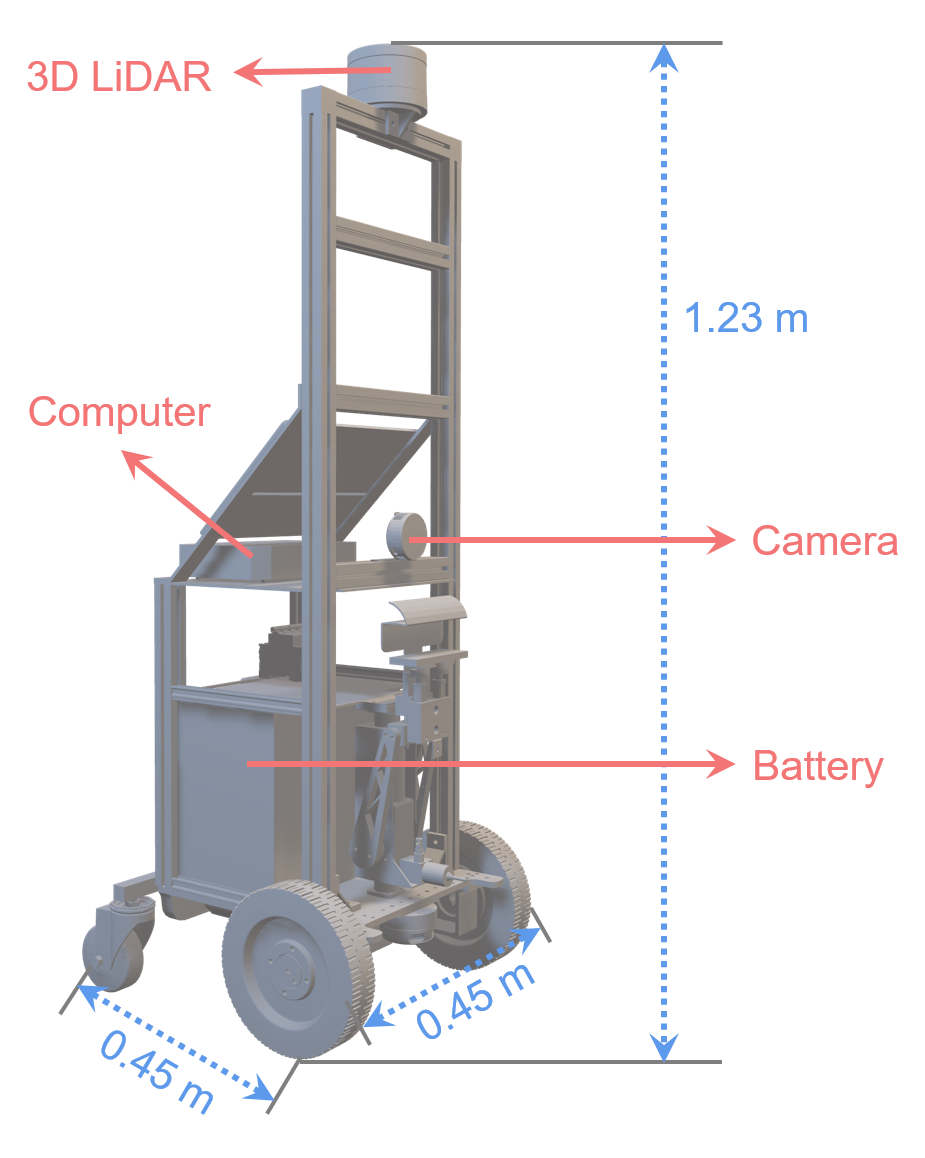}
\caption{Experimental platform for real-world experiment.}
\label{car}
\end{figure}
As shown in Fig. \ref{car}, the robot used in this article measures 0.45 m × 0.45 m × 1.23 m. It is equipped with advanced sensors, including an Ouster OS0-32 LiDAR and a Realsense D435i camera, operating on an onboard computer powered by an i7-1165G7 CPU and an NVIDIA GTX 2060 GPU.

\subsection{Experiment Results}
\begin{figure}[htb]
\centering
    \includegraphics[width=1\columnwidth]{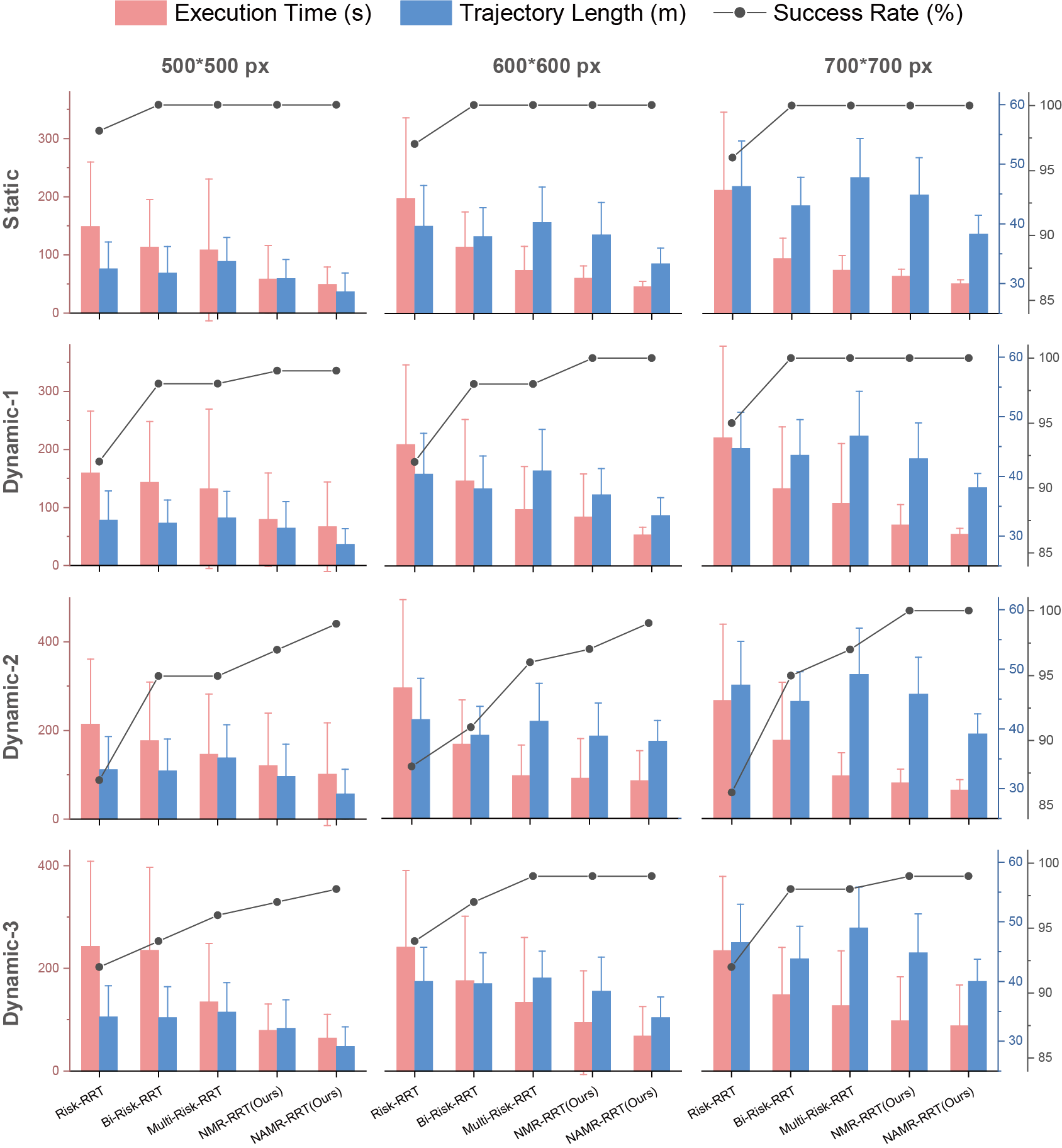}
\caption{Experiment results of five algorithms on Map-1.}
\label{map1}
\end{figure}

\begin{figure}[htb]
\centering
    \includegraphics[width=1\columnwidth]{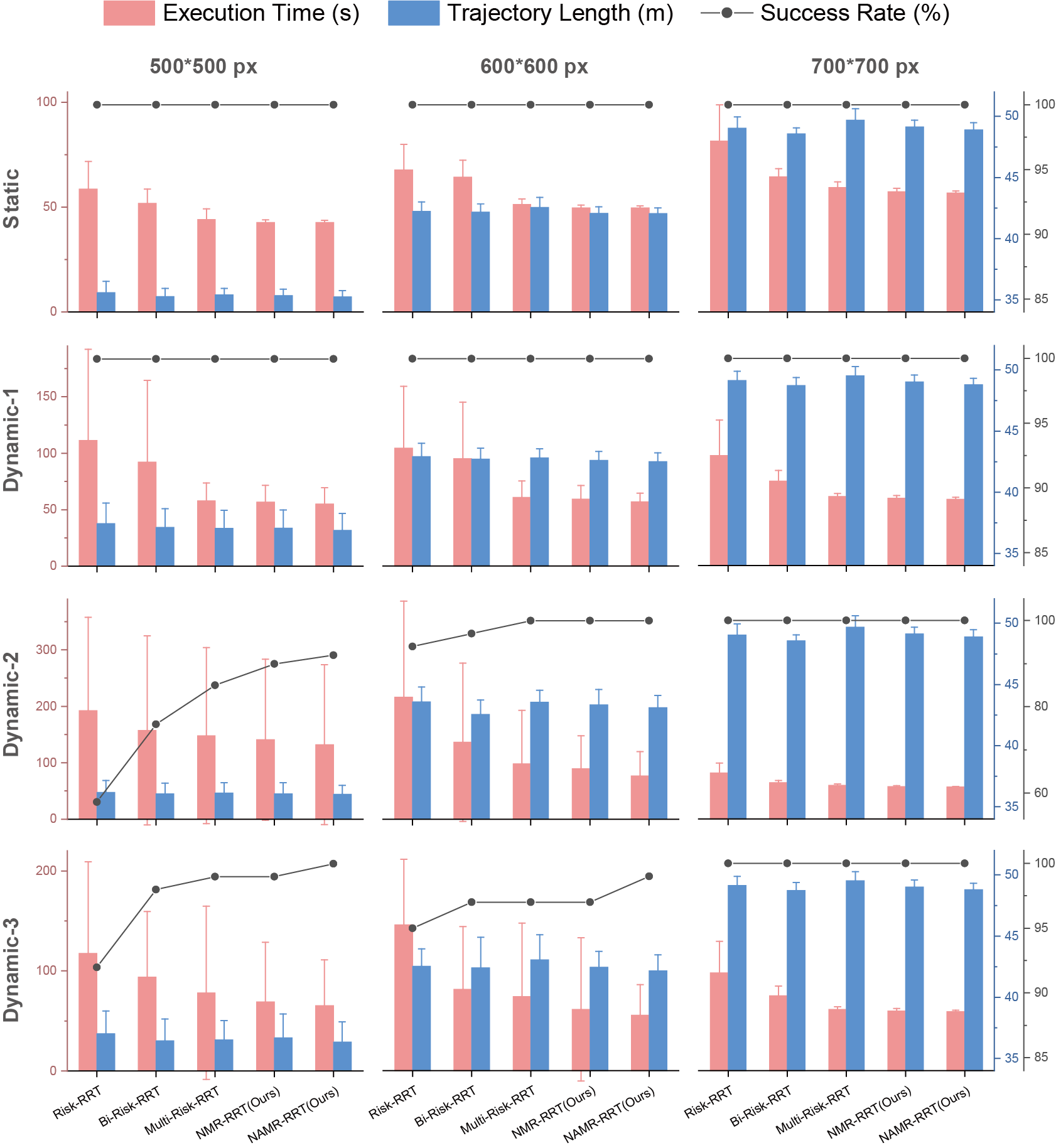}
\caption{Experiment results of five algorithms on Map-2.}
\label{map3}
\end{figure}

\begin{figure}[htb]
\centering
    \includegraphics[width=1\columnwidth]{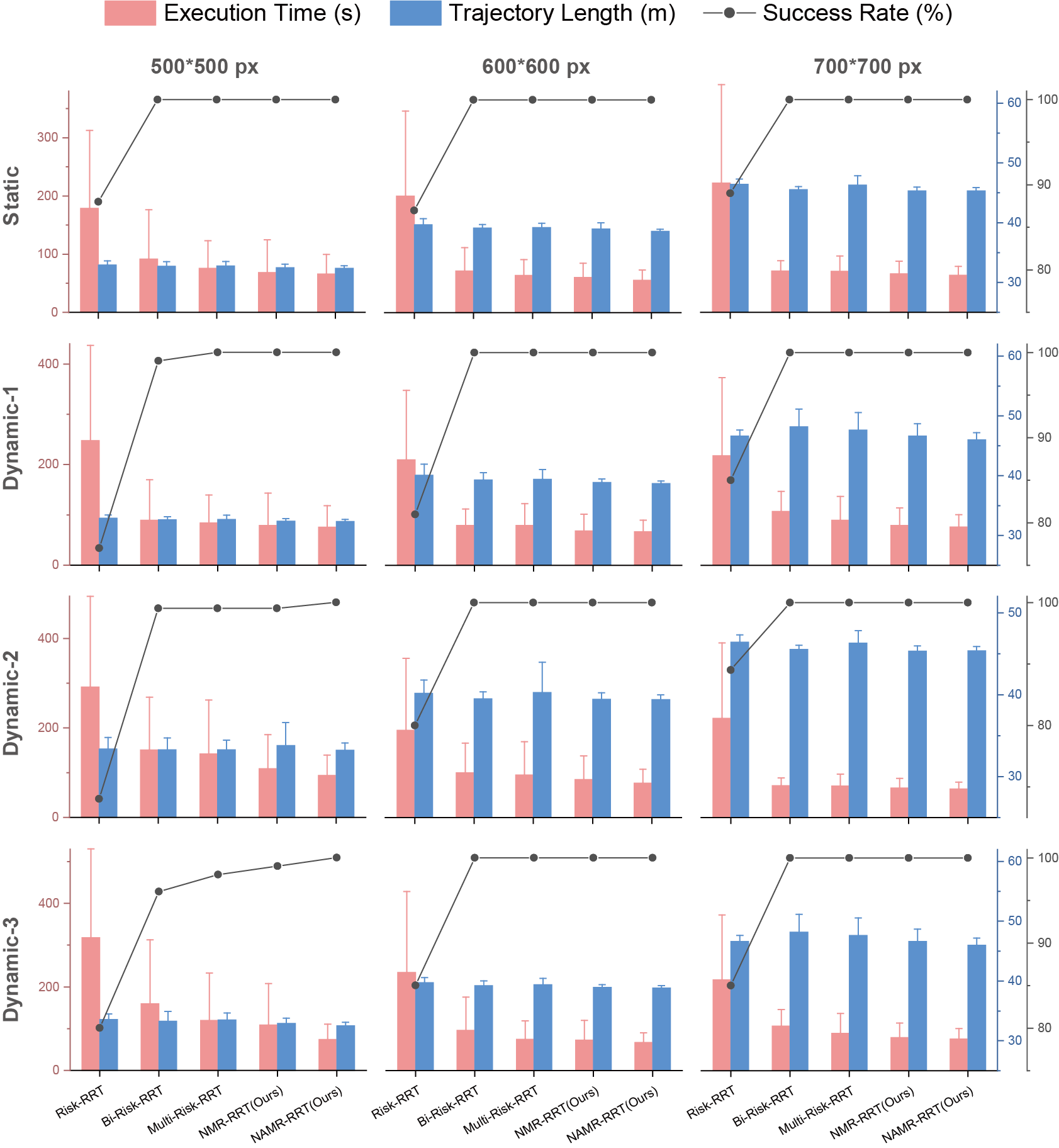}
\caption{Experiment results of five algorithms on Map-3.}
\label{map4}
\end{figure}

\begin{figure}[htb]
\centering
    \includegraphics[width=1\columnwidth]{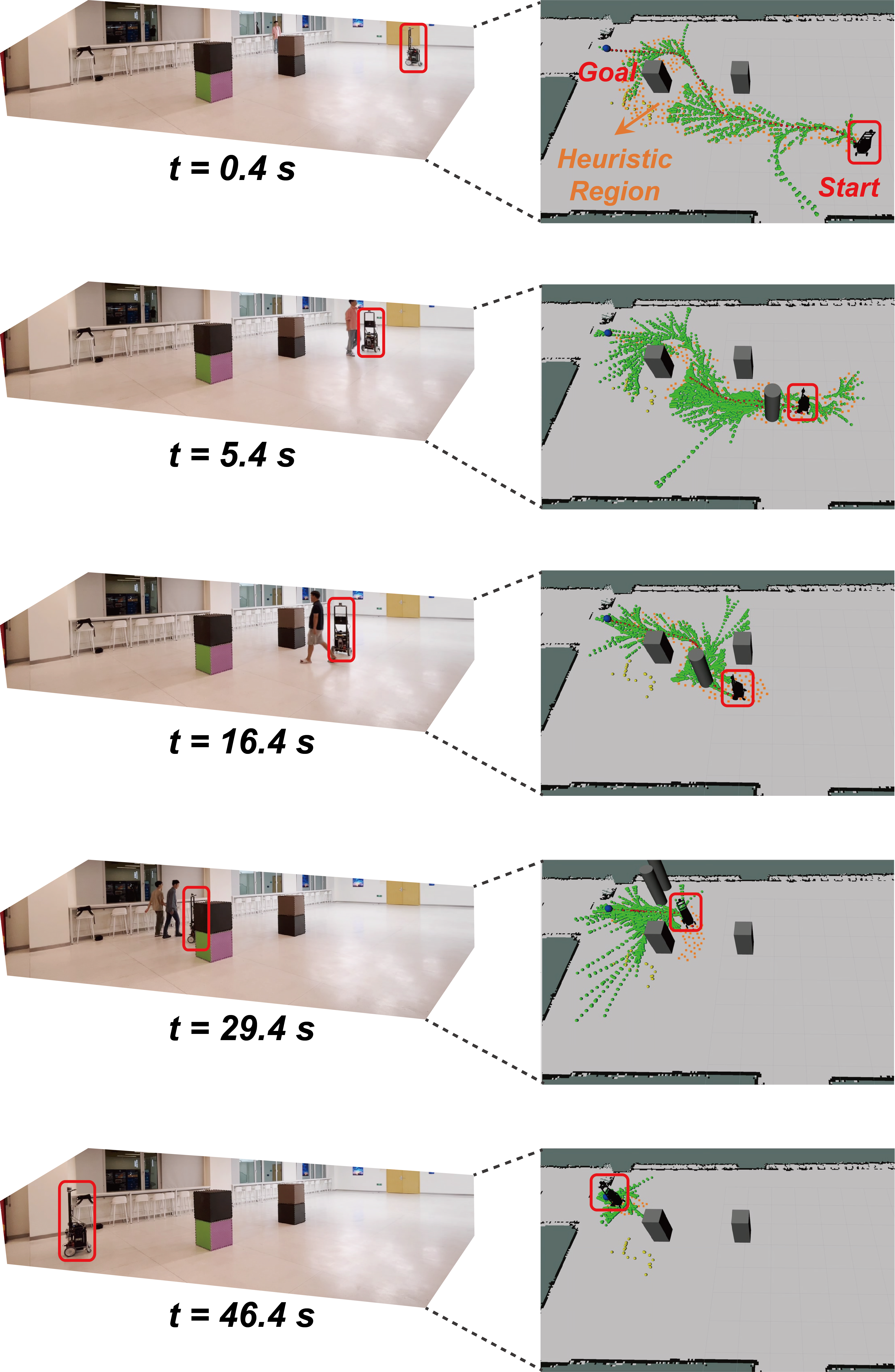}
\caption{Snapshots from the real-world experimental process using NAMR-RRT. Images have been captured at five distinct time points: 0.4 s, 5.4 s, 16.4 s, 29.4 s, and 46.4 s, showing the robot navigating a dynamic indoor environment. On the left is the wide-angle view, and on the right is the visualization view. The red box indicates the robot's position in both views, with the first row showing the robot's start point. In the visualization view, the blue dot represents the robot's goal point, yellow dots indicate the generated heuristic region, green dots represent the exploration, and red dots show the executed trajectory. Besides, the cubes represent static obstacles, while the cylinders indicate moving obstacles.}
\label{real-world}
\end{figure}

Fig. \ref{map1}, Fig. \ref{map3}, and Fig. \ref{map4} present the evaluation results of five motion planning algorithms (Risk-RRT, Bi-Risk-RRT, Multi-Risk-RRT, NMR-RRT, and NAMR-RRT) across three simulation maps (Map-1, Map-2, and Map-3) under both static and dynamic environments (Dynamic-1, Dynamic-2, and Dynamic-3). The results are evaluated using three metrics: Execution Time (seconds, in red), Trajectory Length (meters, in blue), and Success Rate (percentage, in black), with the results averaged from 100  independent runs. The standard deviation for Execution Time and Trajectory Length is indicated by error bars, showing the variability in performance. Additionally, results are presented for varying map sizes (500 × 500, 600 × 600, and 700 × 700 pixels) with a resolution of 0.054, where only the dimensions are scaled while preserving the original map structure. It allows for a comparative analysis of algorithm performance across different map scales. Due to failure cases, the actual mean and standard deviation might be higher than the values shown in the figures. However, this does not impact the overall conclusions of the experiments. As seen in the result figures, even when the baseline algorithms have a lower Success Rate compared to our proposed algorithms, their performance in terms of Execution Time and Trajectory Length remains inferior.

\subsubsection{Execution Time} Across all maps and dynamic environments, NAMR-RRT consistently demonstrates the shortest execution time, with NMR-RRT following closely. Multi-Risk-RRT reduces execution time compared to Risk-RRT and Bi-Risk-RRT, though it still falls behind the neural network-based algorithms. While Bi-Risk-RRT offers better performance than Risk-RRT by leveraging bi-directional search, both algorithms struggle with longer execution time, especially in complex environments. It reflects the limitations of their exploration strategies, as Risk-RRT’s unidirectional search leads to slower planning. At the same time, Bi-Risk-RRT, despite its improvements, remains less efficient than multi-directional and neural network-based approaches.
\subsubsection{Trajectory Length} NAMR-RRT consistently achieves the shortest trajectory lengths across all conditions, closely followed by NMR-RRT, producing slightly longer trajectories. Although multi-directional and bi-directional search strategies reduce execution time, they do not offer an advantage in trajectory length. Consequently, algorithms ranging from unidirectional to bi-directional and multi-directional searches remain less effective in optimizing trajectory length, especially when compared to neural network-based approaches.
\subsubsection{Success Rate} NAMR-RRT and NMR-RRT consistently maintain high success rates across all maps, with NAMR-RRT typically achieving the best performance. Multi-Risk-RRT also demonstrates competitive success rates, significantly outperforming both Bi-Risk-RRT and Risk-RRT. While Bi-Risk-RRT exhibits better success rates than Risk-RRT due to its bi-directional search strategy, both algorithms struggle in environments with higher dynamic complexity.
\subsubsection{Standard Deviation} NAMR-RRT consistently shows the smallest standard deviations across all metrics, reflecting stable and consistent performance. NMR-RRT also exhibits relatively low variability, though higher than NAMR-RRT. In contrast, Risk-RRT, Bi-Risk-RRT, and Multi-Risk-RRT tend to have larger standard deviations, particularly in execution time, indicating greater fluctuations in performance.

In addition to the simulation experiments, we also conduct real-world experiments, as shown in Fig. \ref{real-world}, where a mobile robot uses the NAMR-RRT algorithm to navigate an indoor environment with static obstacles and moving humans. The robot starts at a designated point and plans a trajectory to the goal, dynamically adjusting as pedestrians enter its trajectory. The robot utilizes the heuristic region (represented by the yellow dots in the right-side images) to narrow the search space and generate an efficient trajectory. The heuristic region continuously updates as the robot moves, effectively guiding it towards the goal. The experiment demonstrates the algorithm's ability to handle static and dynamic elements in real-time, ensuring efficient and robust navigation. Please refer to the website link\footnote{\href{https://sites.google.com/view/namr-rrt}{https://sites.google.com/view/namr-rrt}} for a comprehensive view of the experiment process.
\section{Discussion}
\begin{figure*}[htb]
\centering
    \includegraphics[width=2\columnwidth]{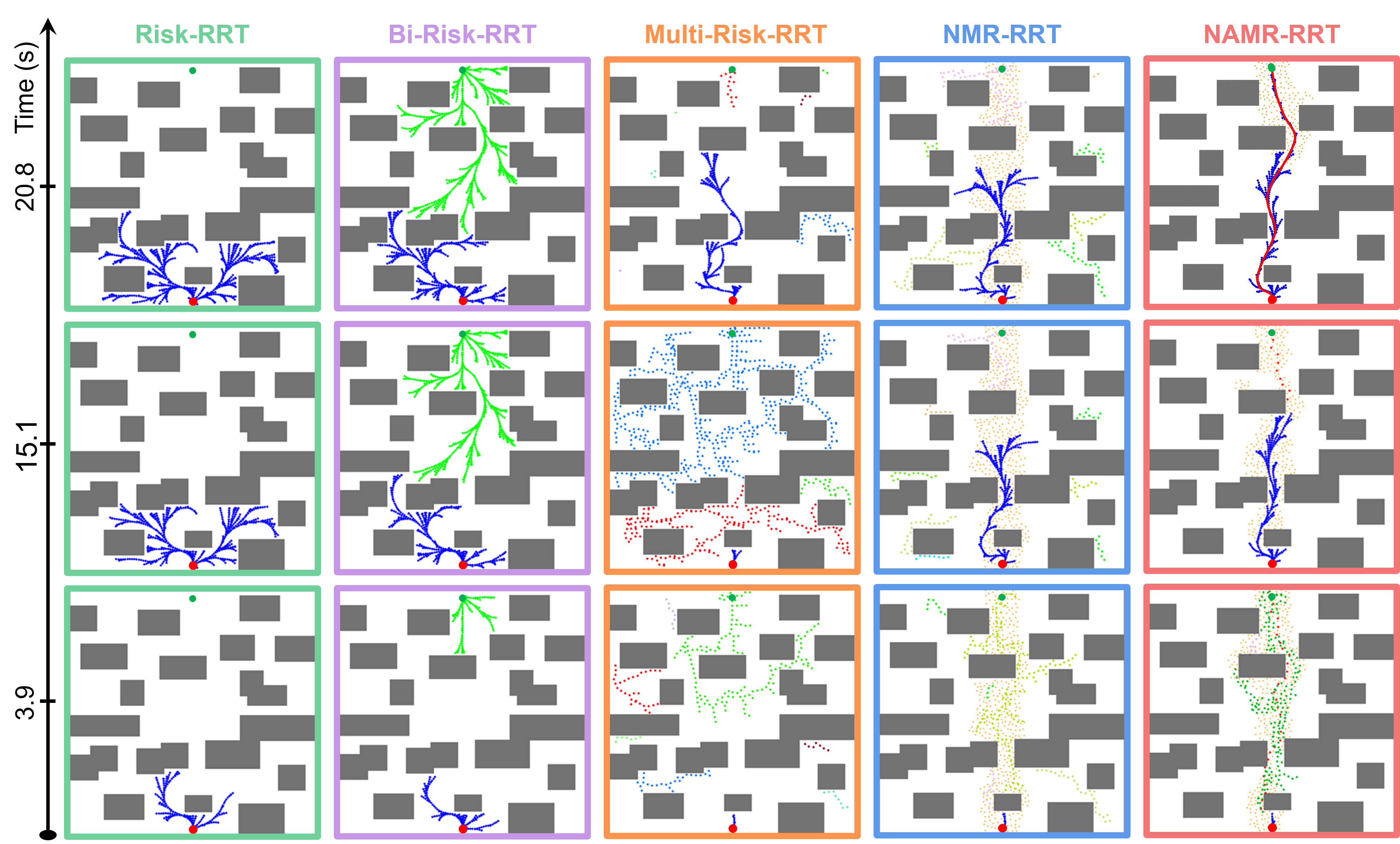}
\caption{Comparison of the search processes of five algorithms. The first column illustrates the performance of Risk-RRT (green), the second column represents Bi-Risk-RRT (purple), the third column shows Multi-Risk-RRT (orange), the fourth column displays NMR-RRT (blue), and the fifth column presents NAMR-RRT (pink). The red and green dots indicate the start and goal points, respectively. The algorithms are evaluated at three distinct time points: t = 3.9 s, t = 15.1 s, and t = 20.8 s.}
\label{compare}
\end{figure*}
This section analyzes the key factors behind the superior performance of NAMR-RRT. Two main aspects drive these improvements: neural network-generated heuristic regions and dynamic updates to the heuristic region and sampling rate. The neural network-generated heuristic region guides the search to the promising areas, reducing unnecessary exploration and improving efficiency. Additionally, dynamic updates allow NAMR-RRT to adapt continuously to changing environments, optimizing trajectory length. The following will discuss the performance of baseline algorithms, the advantages of neural network guidance, and the impact of dynamic updates in NAMR-RRT.

\subsubsection{Performance of Baseline Algorithms} Risk-RRT lacks heuristic guidance, resulting in a slow planning process and a lower-quality trajectory. As shown in Fig. \ref{compare}, the algorithm expands inefficiently, leading to high execution time and long trajectory lengths. Bi-Risk-RRT improves efficiency through bi-directional growth, where two trees grow from the start and goal points, respectively. Once the trees meet, the tree from the goal can guide the tree from the start. However, in complex environments, the meeting time is uncertain or delayed, meaning that Bi-Risk-RRT behaves similarly to Risk-RRT without notable improvements in trajectory quality or planning efficiency for a long time.
Multi-Risk-RRT introduces multi-directional exploration, significantly improving success rate and execution time. Extending bi-directional exploration into a multi-directional approach enhances the rootTree's ability to leverage heuristic information. The growth of multiple subTrees increases the chances of meeting the rootTree, even in complex environments. However, this process remains uncertain, and the search often spans the entire space, resulting in a substantial amount of ineffective exploration. As shown in Fig. \ref{compare}, Multi-Risk-RRT often explores irrelevant space, resulting in a long trajectory. Although the execution time is shortened, the random expansion pattern does not guarantee the quality of the trajectory.

\subsubsection{Advantages of Neural Network-based Algorithms} Our proposed algorithms, NMR-RRT and NAMR-RRT, show clear advantages through the neural network-generated heuristic region. This region enables the algorithms to concentrate their search in promising areas, minimizing unnecessary exploration and significantly improving planning efficiency. As seen in the experiment results and the visualized search process in Fig. \ref{compare}, NMR-RRT benefits from this neural network-generated heuristic region, performing better than the baseline algorithms. However, while NMR-RRT demonstrates a more focused search, it still exhibits scattered subTrees growth. The fixed heuristic region and sampling rate limit the ability to refine the search process. As a result, although NMR-RRT outperforms traditional methods, its motion planning performance falls short of NAMR-RRT, which benefits from dynamic updating capabilities.

\subsubsection{Impact of Adaptive Updates in NAMR-RRT} The introduction of adaptive updates in NAMR-RRT sets it apart from all other algorithms. NAMR-RRT consistently outperforms the others in all metrics by continuously updating the heuristic region and sampling rate. The adaptive updates allow the algorithm to focus more precisely on promising regions, minimizing unnecessary exploration and resulting in shorter trajectory length, faster execution time, and higher success rate. Fig. \ref{compare} demonstrates this, as NAMR-RRT produces a clean and efficient search tree, progressing directly towards the goal with minimal branching or deviation. This efficiency is further supported by the lower standard deviation across all metrics, indicating robust and stable performance, even in complex environments. The contrast between NMR-RRT and NAMR-RRT highlights the critical importance of adaptive updates. While NMR-RRT performs well with a static heuristic, NAMR-RRT's ability to dynamically adjust its focus leads to far more efficient motion planning. The visual comparison shows how NAMR-RRT avoids the excessive branching in the baseline algorithms and converges rapidly on the high-quality trajectory.

In conclusion, the experiment results validate the effectiveness of neural network-based, adaptively updated algorithms in solving motion planning problems, especially in dynamic and complex environments. NAMR-RRT's superior performance across all metrics confirms the advantages of combining neural network guidance with adaptive updates, offering a robust and reliable solution for navigation tasks. 
\section{Conclusions and Future Work}\label{future}
This article presents NAMR-RRT, a neural adaptive motion planning algorithm for efficient navigation in dynamic environments. NAMR-RRT efficiently focuses on promising areas, minimizing unnecessary exploration and ensuring robust performance even under challenging conditions.
The experiment results demonstrate that NAMR-RRT consistently outperforms traditional methods and NMR-RRT, a neural network-based method, which operates with a fixed heuristic region and sampling rate. NAMR-RRT achieves superior results in execution time, trajectory length, and success rate. These findings emphasize the importance of integrating neural network-generated heuristic regions with adaptive updates to the heuristic regions and sampling rate, offering a practical and effective solution for navigation in dynamic and complex environments. 
Future work will investigate advanced neural network models for generating more refined heuristic regions while integrating imperative learning to enable the algorithm to learn from past experiences and enhance the search process.
\bibliographystyle{IEEEtran} 
\bibliography{refs} 

\begin{thebibliography}{10}
\providecommand{\url}[1]{#1}
\csname url@samestyle\endcsname
\providecommand{\newblock}{\relax}
\providecommand{\bibinfo}[2]{#2}
\providecommand{\BIBentrySTDinterwordspacing}{\spaceskip=0pt\relax}
\providecommand{\BIBentryALTinterwordstretchfactor}{4}
\providecommand{\BIBentryALTinterwordspacing}{\spaceskip=\fontdimen2\font plus
\BIBentryALTinterwordstretchfactor\fontdimen3\font minus \fontdimen4\font\relax}
\providecommand{\BIBforeignlanguage}[2]{{%
\expandafter\ifx\csname l@#1\endcsname\relax
\typeout{** WARNING: IEEEtran.bst: No hyphenation pattern has been}%
\typeout{** loaded for the language `#1'. Using the pattern for}%
\typeout{** the default language instead.}%
\else
\language=\csname l@#1\endcsname
\fi
#2}}
\providecommand{\BIBdecl}{\relax}
\BIBdecl

\bibitem{agv}
F.~Pratissoli, R.~Brugioni, N.~Battilani, and L.~Sabattini, ``Hierarchical traffic management of multi-agv systems with deadlock prevention applied to industrial environments,'' \emph{IEEE Transactions on Automation Science and Engineering}, 2023.

\bibitem{cleaning}
M.~M. Rayguru, S.~Roy, L.~Yi, M.~R. Elara, and S.~Baldi, ``Introducing switched adaptive control for self-reconfigurable mobile cleaning robots,'' \emph{IEEE Transactions on Automation Science and Engineering}, vol.~21, no.~2, pp. 2051--2062, 2023.

\bibitem{A*}
P.~E. Hart, N.~J. Nilsson, and B.~Raphael, ``A formal basis for the heuristic determination of minimum cost paths,'' \emph{IEEE transactions on Systems Science and Cybernetics}, vol.~4, no.~2, pp. 100--107, 1968.

\bibitem{Dijkstra}
E.~W. Dijkstra, ``A note on two problems in connexion with graphs,'' in \emph{Edsger Wybe Dijkstra: his life, work, and legacy}, 2022, pp. 287--290.

\bibitem{apf}
O.~Khatib, ``Real-time obstacle avoidance for manipulators and mobile robots,'' \emph{The international journal of robotics research}, vol.~5, no.~1, pp. 90--98, 1986.

\bibitem{smothness}
Y.~Guo, D.~Yao, B.~Li, Z.~He, H.~Gao, and L.~Li, ``Trajectory planning for an autonomous vehicle in spatially constrained environments,'' \emph{IEEE Transactions on Intelligent Transportation Systems}, vol.~23, no.~10, pp. 18\,326--18\,336, 2022.

\bibitem{traj-generate}
J.~Ziegler, P.~Bender, M.~Schreiber, H.~Lategahn, T.~Strauss, C.~Stiller, T.~Dang, U.~Franke, N.~Appenrodt, C.~G. Keller \emph{et~al.}, ``Making bertha drive—an autonomous journey on a historic route,'' \emph{IEEE Intelligent transportation systems magazine}, vol.~6, no.~2, pp. 8--20, 2014.

\bibitem{rl}
J.~Yan, K.~You, W.~Cao, X.~Yang, and X.~Guan, ``Binocular vision-based motion planning of an auv: A deep reinforcement learning approach,'' \emph{IEEE Transactions on Intelligent Vehicles}, 2023.

\bibitem{PRM}
L.~E. Kavraki, P.~Svestka, J.-C. Latombe, and M.~H. Overmars, ``Probabilistic roadmaps for path planning in high-dimensional configuration spaces,'' \emph{IEEE transactions on Robotics and Automation}, vol.~12, no.~4, pp. 566--580, 1996.

\bibitem{RRT}
S.~M. LaValle and J.~J. Kuffner~Jr, ``Randomized kinodynamic planning,'' \emph{The international journal of robotics research}, vol.~20, no.~5, pp. 378--400, 2001.

\bibitem{rrt-connect}
J.~J. Kuffner and S.~M. LaValle, ``Rrt-connect: An efficient approach to single-query path planning,'' in \emph{Proceedings 2000 ICRA. Millennium Conference. IEEE International Conference on Robotics and Automation. Symposia Proceedings (Cat. No. 00CH37065)}, vol.~2.\hskip 1em plus 0.5em minus 0.4em\relax IEEE, 2000, pp. 995--1001.

\bibitem{rrdt}
T.~Lai, F.~Ramos, and G.~Francis, ``Balancing global exploration and local-connectivity exploitation with rapidly-exploring random disjointed-trees,'' in \emph{2019 International Conference on Robotics and Automation (ICRA)}.\hskip 1em plus 0.5em minus 0.4em\relax IEEE, 2019, pp. 5537--5543.

\bibitem{constrain}
J.-P. Laumond, P.~E. Jacobs, M.~Taix, and R.~M. Murray, ``A motion planner for nonholonomic mobile robots,'' \emph{IEEE Transactions on robotics and automation}, vol.~10, no.~5, pp. 577--593, 1994.

\bibitem{tbvp}
L.~Quartapelle and S.~Rebay, ``Numerical solution of two-point boundary value problems,'' \emph{Journal of computational physics}, vol.~86, no.~2, pp. 314--354, 1990.

\bibitem{b2u}
J.~Wang, W.~Chi, C.~Li, and M.~Q.-H. Meng, ``Efficient robot motion planning using bidirectional-unidirectional rrt extend function,'' \emph{IEEE Transactions on Automation Science and Engineering}, vol.~19, no.~3, pp. 1859--1868, 2021.

\bibitem{mt-rrt}
Z.~Sun, J.~Wang, and M.~Q.-H. Meng, ``Multi-tree guided efficient robot motion planning,'' \emph{Procedia Computer Science}, vol. 209, pp. 31--39, 2022.

\bibitem{risk-rrt}
C.~Fulgenzi, A.~Spalanzani, C.~Laugier, and C.~Tay, ``Risk based motion planning and navigation in uncertain dynamic environment,'' \emph{Research Report}, 2010.

\bibitem{bi-risk-rrt}
H.~Ma, F.~Meng, C.~Ye, J.~Wang, and M.~Q.-H. Meng, ``Bi-risk-rrt based efficient motion planning for autonomous ground vehicles,'' \emph{IEEE Transactions on Intelligent Vehicles}, vol.~7, no.~3, pp. 722--733, 2022.

\bibitem{Multi-Risk-RRT}
Z.~Sun, B.~Lei, P.~Xie, F.~Liu, J.~Gao, Y.~Zhang, and J.~Wang, ``Multi-risk-rrt: An efficient motion planning algorithm for robotic autonomous luggage trolley collection at airports,'' \emph{IEEE Transactions on Intelligent Vehicles}, 2024.

\bibitem{rrt*}
S.~Karaman and E.~Frazzoli, ``Sampling-based algorithms for optimal motion planning,'' \emph{The international journal of robotics research}, vol.~30, no.~7, pp. 846--894, 2011.

\bibitem{Informed_rrt}
J.~D. Gammell, S.~S. Srinivasa, and T.~D. Barfoot, ``Informed rrt*: Optimal sampling-based path planning focused via direct sampling of an admissible ellipsoidal heuristic,'' in \emph{2014 IEEE/RSJ international conference on intelligent robots and systems}.\hskip 1em plus 0.5em minus 0.4em\relax IEEE, 2014, pp. 2997--3004.

\bibitem{bit}
J.~D. Gammell, T.~D. Barfoot, and S.~S. Srinivasa, ``Batch informed trees (bit*): Informed asymptotically optimal anytime search,'' \emph{The International Journal of Robotics Research}, vol.~39, no.~5, pp. 543--567, 2020.

\bibitem{mprrt}
M.~Zucker, J.~Kuffner, and M.~Branicky, ``Multipartite rrts for rapid replanning in dynamic environments,'' in \emph{Proceedings 2007 IEEE International Conference on Robotics and Automation}.\hskip 1em plus 0.5em minus 0.4em\relax IEEE, 2007, pp. 1603--1609.

\bibitem{d*}
A.~Stentz, ``Optimal and efficient path planning for partially-known environments,'' in \emph{Proceedings of the 1994 IEEE international conference on robotics and automation}.\hskip 1em plus 0.5em minus 0.4em\relax IEEE, 1994, pp. 3310--3317.

\bibitem{d*-lite}
S.~Koenig and M.~Likhachev, ``D* lite,'' in \emph{Eighteenth national conference on Artificial intelligence}, 2002, pp. 476--483.

\bibitem{rrtx}
M.~Otte and E.~Frazzoli, ``Rrtx: Asymptotically optimal single-query sampling-based motion planning with quick replanning,'' \emph{The International Journal of Robotics Research}, vol.~35, no.~7, pp. 797--822, 2016.

\bibitem{re-rrt}
K.~Naderi, J.~Rajam{\"a}ki, and P.~H{\"a}m{\"a}l{\"a}inen, ``Rt-rrt* a real-time path planning algorithm based on rrt,'' in \emph{Proceedings of the 8th ACM SIGGRAPH Conference on Motion in Games}, 2015, pp. 113--118.

\bibitem{neural-rrt}
J.~Wang, W.~Chi, C.~Li, C.~Wang, and M.~Q.-H. Meng, ``Neural rrt*: Learning-based optimal path planning,'' \emph{IEEE Transactions on Automation Science and Engineering}, vol.~17, no.~4, pp. 1748--1758, 2020.

\bibitem{neural-informed-rrt}
Z.~Huang, H.~Chen, J.~Pohovey, and K.~Driggs-Campbell, ``Neural informed rrt*: Learning-based path planning with point cloud state representations under admissible ellipsoidal constraints,'' in \emph{2024 IEEE International Conference on Robotics and Automation (ICRA)}.\hskip 1em plus 0.5em minus 0.4em\relax IEEE, 2024, pp. 8742--8748.

\bibitem{tis}
F.~Meng, J.~Liu, H.~Shi, H.~Ma, H.~Ren, and M.~Q.-H. Meng, ``Online time-informed kinodynamic motion planning of nonlinear systems,'' \emph{IEEE Robotics and Automation Letters}, 2024.

\bibitem{gan}
T.~Zhang, J.~Wang, and M.~Q.-H. Meng, ``Generative adversarial network based heuristics for sampling-based path planning,'' \emph{IEEE/CAA Journal of Automatica Sinica}, vol.~9, no.~1, pp. 64--74, 2021.

\bibitem{risk-dtrrt}
W.~Chi, C.~Wang, J.~Wang, and M.~Q.-H. Meng, ``Risk-dtrrt-based optimal motion planning algorithm for mobile robots,'' \emph{IEEE Transactions on Automation Science and Engineering}, vol.~16, no.~3, pp. 1271--1288, 2018.

\bibitem{pointnet++}
C.~R. Qi, L.~Yi, H.~Su, and L.~J. Guibas, ``Pointnet++: Deep hierarchical feature learning on point sets in a metric space,'' \emph{Advances in neural information processing systems}, vol.~30, 2017.

\bibitem{sdg}
L.~Bottou, ``Large-scale machine learning with stochastic gradient descent,'' in \emph{Proceedings of COMPSTAT'2010: 19th International Conference on Computational StatisticsParis France, August 22-27, 2010 Keynote, Invited and Contributed Papers}.\hskip 1em plus 0.5em minus 0.4em\relax Springer, 2010, pp. 177--186.

\bibitem{ucy}
A.~Lerner, Y.~Chrysanthou, and D.~Lischinski, ``Crowds by example,'' in \emph{Computer graphics forum}, vol.~26, no.~3.\hskip 1em plus 0.5em minus 0.4em\relax Wiley Online Library, 2007, pp. 655--664.

\end{thebibliography}

\vspace{-10 mm}
\begin{IEEEbiography}
[{\includegraphics[width=1in,height=1.25in,clip,keepaspectratio]{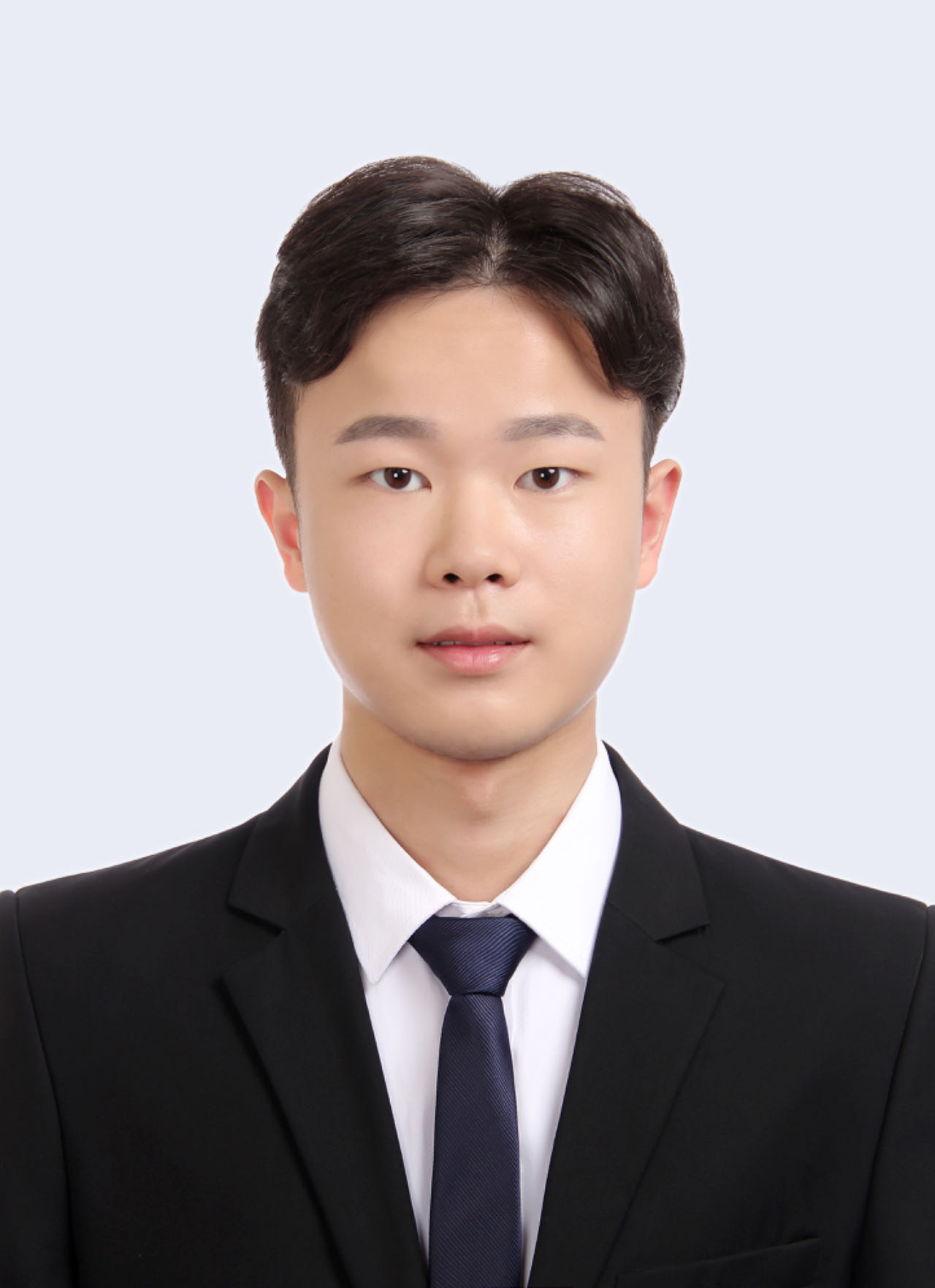}}] 
{Zhirui Sun} received the B.E. degree in information engineering from the Department of Electronic and Electrical Engineering, Southern University of Science and Technology, Shenzhen, China, in 2019. He is currently pursuing the Ph.D. degree with the Department of Electronic and Electrical Engineering, Southern University of Science and Technology, Shenzhen, China. His research interests include robot perception and motion planning.
\end{IEEEbiography}
\vspace{-10 mm}

\begin{IEEEbiography}
[{\includegraphics[width=1in,height=1.25in,clip,keepaspectratio]{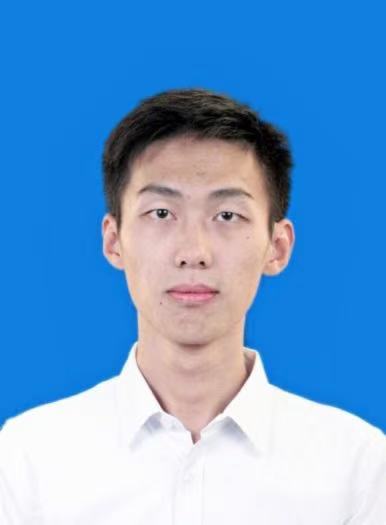}}] 
{Bingyi Xia} received the B.E. degree in microelectronics science and engineering from the Southern University of Science and Technology, Shenzhen, China, in 2020. He is currently pursuing the Ph.D. degree with the Department of Electronic and Electrical Engineering, Southern University of Science and Technology, Shenzhen, China. His research interests include robot motion planning.
\end{IEEEbiography}
\vspace{-10 mm}

\begin{IEEEbiography}
[{\includegraphics[width=1in,height=1.25in,clip,keepaspectratio]{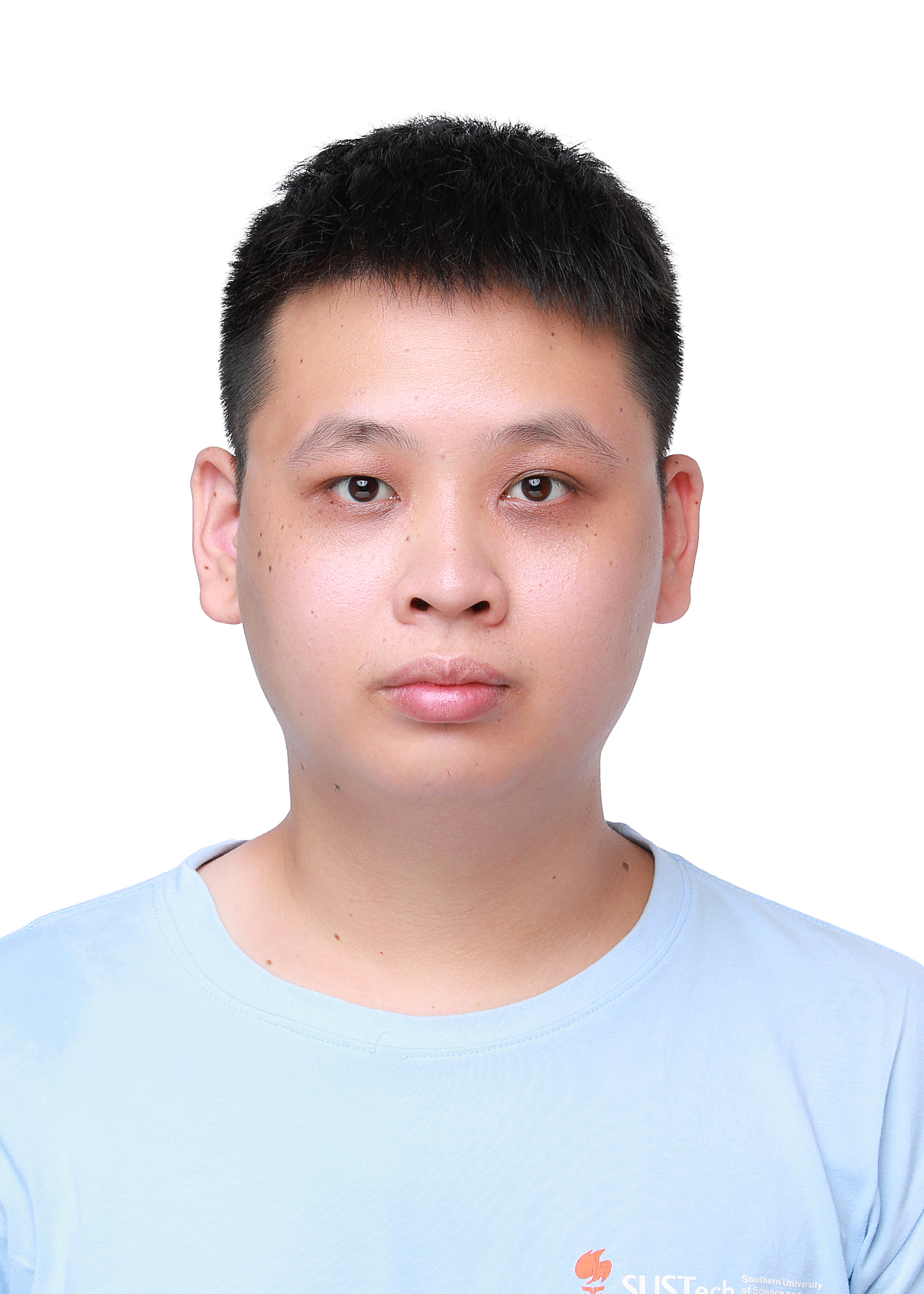}}] 
{Peijia Xie} received the B.E. degree in Electronic Information Engineering from the School of Physics and Telecommunications Engineering, South China Normal University, Guangzhou, China, in 2022. He is currently pursuing the Master degree with the Department of Electronic and Electrical Engineering, Southern University of Science and Technology, Shenzhen, China. His research interests include autonomous driving.
\end{IEEEbiography}
\vspace{-10 mm}

\begin{IEEEbiography}
[{\includegraphics[width=1in,height=1.25in,clip,keepaspectratio]{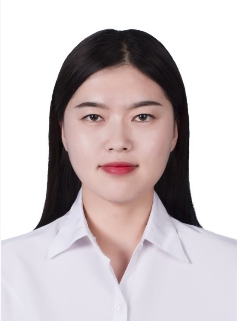}}] 
{Xiaoxiao Li} received the B.E. and M.S. degree in Computer science and technology from Qufu Normal University, China, in 2016 and 2019, respectively. She is currently pursuing the Ph.D. degree with the Harbin Institute of Technology, Shenzhen, China. Her current research interests include intelligent optimization, motion planning, deep learning and robotic control.
\end{IEEEbiography}
\vspace{-10 mm}

\begin{IEEEbiography}
[{\includegraphics[width=1in,height=1.25in,clip,keepaspectratio]{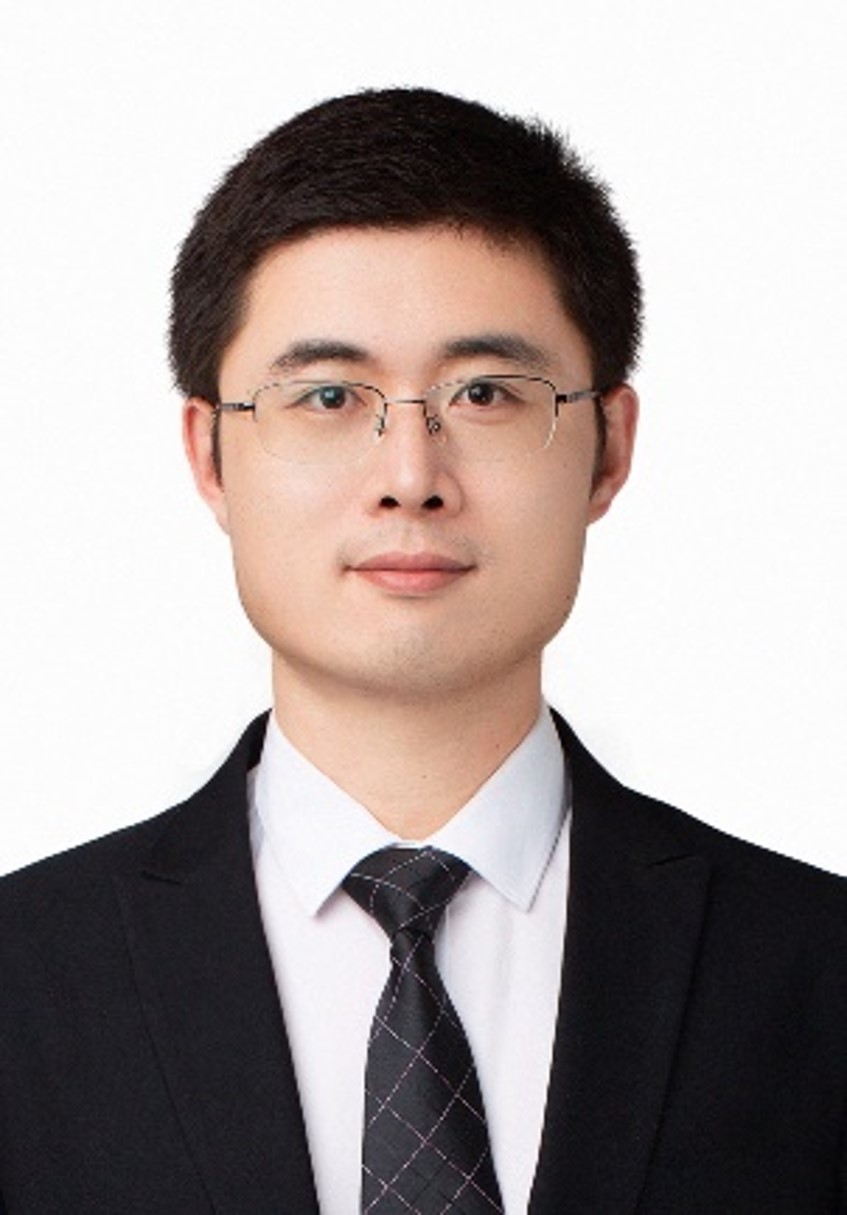}}] 
{Jiankun Wang} (Senior Member, IEEE) received the B.E. degree in automation from Shandong University, Jinan, China, in 2015, and the Ph.D. degree from the Department of Electronic Engineering, The Chinese University of Hong Kong, Hong Kong, in 2019.

During his Ph.D. degree, he spent six months with Stanford University, Stanford, CA, USA, as a Visiting Student Scholar, supervised by Prof. Oussama Khatib. He is currently an Assistant Professor with the Department of Electronic and Electrical Engineering, Southern University of Science and Technology, Shenzhen, China. His current research interests include motion planning and control, human–robot interaction, and machine learning in robotics.

Currently, he serves as the associate editor of IEEE Transactions on Automation Science and Engineering, IEEE Transactions on Intelligent Vehicles, IEEE Robotics and Automation Letters, International Journal of Robotics and Automation, and Biomimetic Intelligence and Robotics.
\end{IEEEbiography}

\end{document}